\documentclass[final,5p,times,twocolumn]{elsarticle}

\usepackage{graphicx}
\usepackage{amsmath,amssymb}
\usepackage{xcolor}
\usepackage{caption}
\usepackage{subcaption}  
\usepackage{url}
\usepackage[hidelinks]{hyperref}
\usepackage{booktabs}
\usepackage{multirow}
\usepackage{algorithm}
\usepackage{algorithmic}
\usepackage{footmisc}
\usepackage{makecell}

\journal{ISPRS Journal of Photogrammetry and Remote Sensing}

\begin{document}

\begin{frontmatter}
	
	\title{Cross-Modal Bidirectional Interaction Model for Referring Remote Sensing Image Segmentation}
	
	\author{
		Zhe Dong\textsuperscript{a,*}, 
		Yuzhe Sun\textsuperscript{a,*}, 
		Tianzhu Liu\textsuperscript{a}, 
		Wangmeng Zuo\textsuperscript{b,c}, 
		Yanfeng Gu\textsuperscript{a,\dag}
	}
	
	\address{
		\textsuperscript{a}School of Electronics and Information Engineering, Harbin Institute of Technology, Harbin 150001, China \\
		\textsuperscript{b}Faculty of Computing, Harbin Institute of Technology, Harbin 150001, China \\
		\textsuperscript{c}Peng Cheng Laboratory, Shenzhen 518055, China
	}
	
\begin{abstract}

Referring remote sensing image segmentation (RRSIS) aims to generate pixel-level masks of target objects in remote sensing images based on natural language expressions. This task is inherently challenging due to the complex geospatial relationships and varying object scales in remote sensing scenarios. To address the aforementioned challenges, a novel RRSIS framework is proposed, termed the cross-modal bidirectional interaction model (CroBIM). Specifically, a context-aware prompt modulation (CAPM) module is designed to integrate spatial positional relationships and task-specific knowledge into the linguistic features, thereby enhancing the ability to capture the target object. Additionally, a language-guided feature aggregation (LGFA) module is introduced  to incorporate linguistic cues into multi-scale visual features, with an attention deficit compensation mechanism being applied to ensure robust alignment across object scales. Finally, a mutual-interaction decoder (MID) is developed to align vision and language features through iterative bidirectional refinement, enabling precise segmentation even in complex geospatial contexts. Moreover we also construct RISBench, a new large-scale benchmark dataset comprising 52,472 image-language-label triplets. Extensive benchmarking on RISBench and two other prevalent datasets demonstrates the superior performance of the proposed CroBIM over existing state-of-the-art (SOTA) methods. The source code for CroBIM and the RISBench dataset will be publicly available at \href{https://github.com/HIT-SIRS/CroBIM}{https://github.com/HIT-SIRS/CroBIM}.
\end{abstract}

\begin{keyword}
Vision and language \sep Referring remote sensing image segmentation (RRSIS) \sep Cross-modal.
\end{keyword}

\end{frontmatter}
\renewcommand{\thefootnote}{}  
\footnotetext{\textsuperscript{*} These authors contributed equally to this work.}
\footnotetext{\textsuperscript{\dag} Corresponding author. E-mail address: guyf@hit.edu.cn (Y. Gu).}
\renewcommand{\thefootnote}{\arabic{footnote}}  

\section{Introduction}

Deep learning has become a cornerstone for remote sensing applications\cite{10151935,10255654}, enabling tasks such as semantic segmentation \cite{dong2023distilling}, object detection\cite{cai2024poly}, and change detection\cite{chen2021remote}. However, these methods often focus on visual comprehension, neglecting the modeling of object relationships and deeper semantic understanding.

Recent advancements in large language models (LLMs) have spurred research into vision-language models (VLMs), enabling tasks such as image captioning\cite{zia2022transforming, zhao2021high}, image-text retrieval\cite{al2022multilanguage, yuan2022remote}, text-based remote sensing image generation\cite{xu2023txt2img, zhao2021text}, and visual question answering\cite{yuan2022easy, zheng2021mutual}. However, the task of referring remote sensing image segmentation (RRSIS) remains relatively unexplored.

\begin{figure}[tbp]
	\begin{center}
		\centerline{\includegraphics[width=1\linewidth]{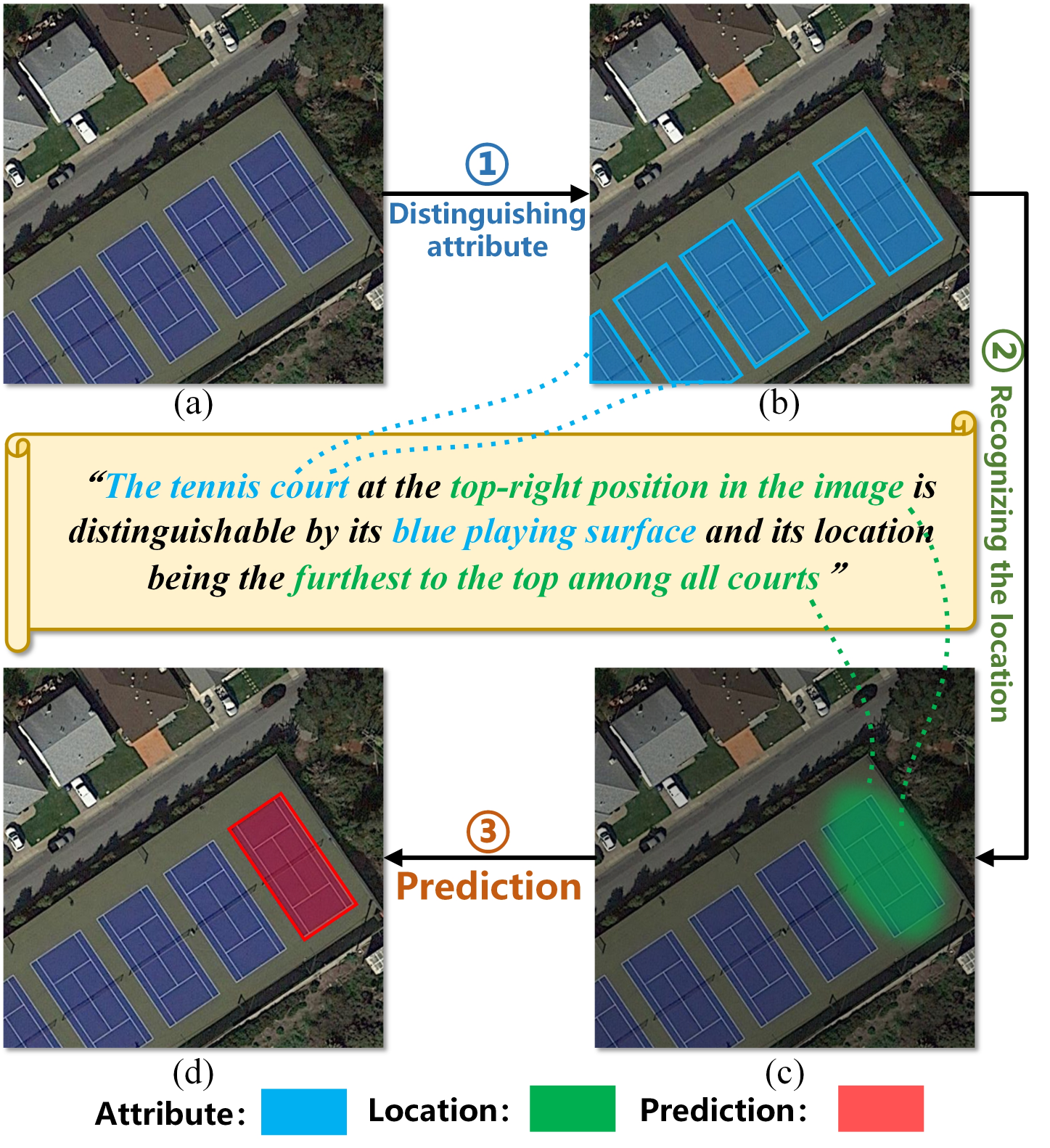}}
		\caption{Illustration of the RRSIS task. (a) The input consists of a referring expression and an image. (b) The model first identifies all candidate objects described in the expression based on information such as category, color, and shape (e.g., `tennis court' and `blue playing surface'). (c) After identifying all potential candidate objects that match the input expression, additional information such as position and size (e.g., `top-right position', `furthest to the top among all courts') is utilized to highlight the target object. (d) Through relation-aware reasoning, the final segmentation mask of the predicted object is obtained.}\label{sketch}
	\end{center}
\end{figure}

As illustrated in Fig.~\ref{sketch}, given remote sensing images and language expressions, RRSIS aims to provide pixel-level masks for specific regions or objects based on the content of the images and expressions. Its core principle is to achieve precise object localization and segmentation by matching textual descriptions with image content. RRSIS breaks the boundaries of traditional semantic understanding of remote sensing data, enabling non-expert users to retrieve objects in remote sensing images through human-computer interaction. It has broad application prospects in land use analysis\cite{mohanrajan2020survey}, search and rescue operations\cite{wang2018deep}, environmental monitoring\cite{yuan2020deep}, military intelligence generation\cite{zhang2022progress}, agricultural production\cite{weiss2020remote}, and urban planning\cite{wellmann2020remote}.

Although referring image segmentation in natural scenarios has made some progress, research on RRSIS is still in its infancy. Yuan \textit{et al.}\cite{yuan2024rrsis} first introduced the concept of the RRSIS task and proposed a language-guided cross-scale enhancement (LGCE) module based on the language-aware vision Transformer (LAVT)\cite{yang2022lavt} to improve segmentation performance for small and sparsely distributed objects. Furthermore, rotated multi-scale interaction network (RMSIN)\cite{liu2024rotated} is designed to address the prevalent challenges of complex scales and orientations in RRSIS. To manage cross-scale fine-grained information, the intra-scale interaction module (IIM) and cross-scale interaction module (CIM) are developed. Additionally, adaptive rotated convolution (ARC) is introduced to enhance the model's robustness to rotational variations. The aforementioned methods rely solely on jointly embedding linguistic features during visual encoding to perceive relevant linguistic context at each spatial location. Although these approaches have achieved satisfactory performance, the interrelation and alignment of visual and linguistic features across multiple levels of the encoding process have not yet been thoroughly explored. 

\begin{figure}[tbp]
	\begin{center}
		\centerline{\includegraphics[width=1\linewidth]{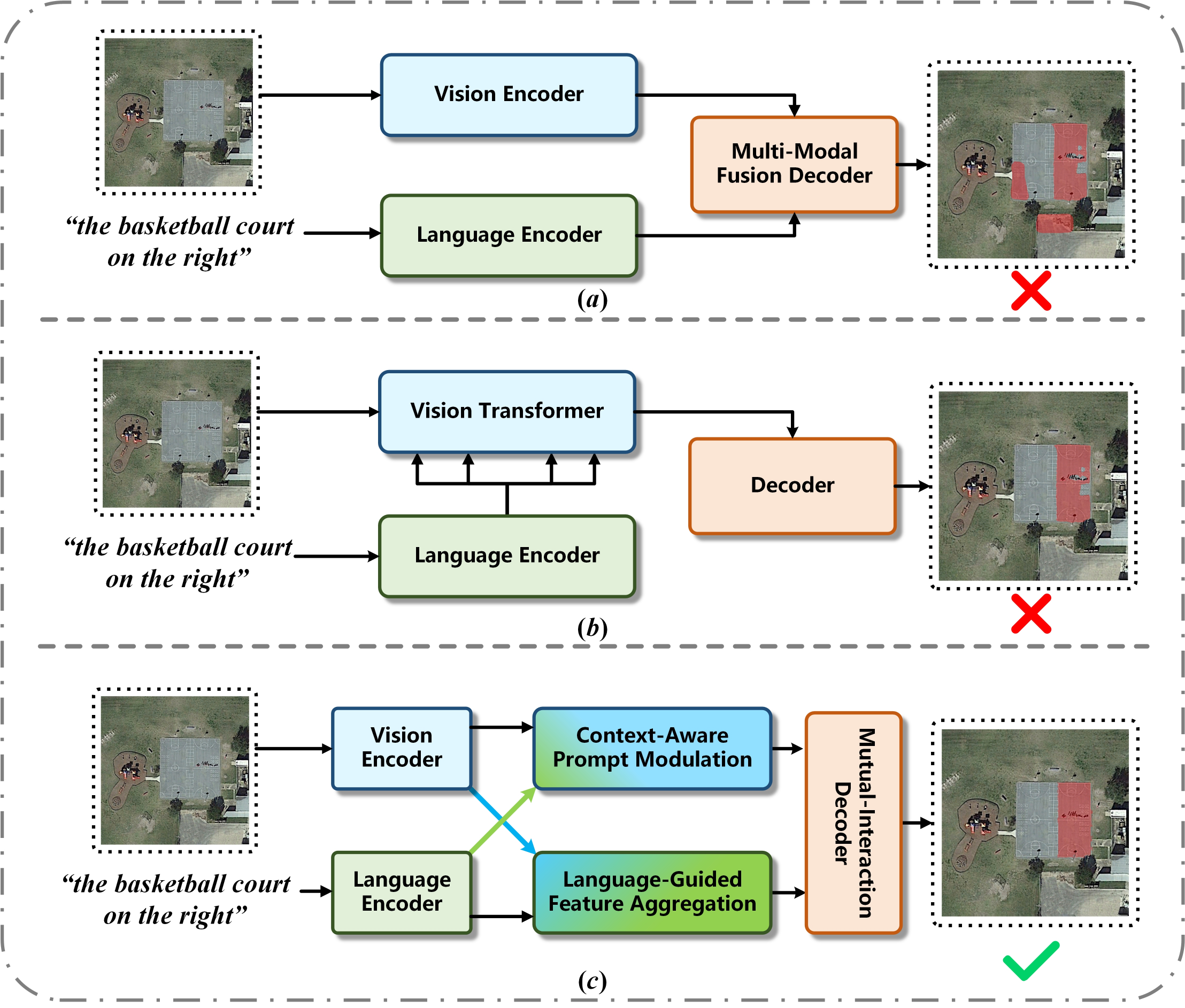}}
		\caption{Conceptual comparison of RRSIS frameworks: (a) cross-modal feature fusion during decoding, (b) directly integrating linguistic information into visual features, and (c) our cross-modal bidirectional interaction model (CroBIM) model.}\label{comparison}
	\end{center}
\end{figure}

Specifically, as shown in Fig.~\ref{comparison}(a) and Fig.~\ref{comparison}(b), failing to consider the underlying correlations between linguistic and visual information and merely fusing cross-modal heterogeneous features at different stages can lead to attention drift, resulting in a mismatch between visual features and the regions described in the query expression. Moreover, compared to natural scene images, the diversity of remote sensing data and the complex geospatial relationships embedded in the corresponding expressions present significant challenges for accurately locating and segmenting the target regions.

In this paper, we introduce a novel cross-modal bidirectional interaction model (CroBIM) for the RRSIS task, addressing the previously identified challenges. As depicted in Fig.~\ref{comparison}(c), the essence of CroBIM lies in its capability to facilitate bidirectional interaction and correlation between visual and linguistic features throughout both the encoding and decoding phases. This enables precise visual-linguistic alignment during the prediction stage. Specifically, the context-aware prompt modulation (CAPM) module is introduced to enhance the text feature encoding process by incorporating multi-scale visual contextual information via learnable prompts. This integration enables the model to effectively perceive the spatial structure and relative positioning of target objects described in the referring expressions. Additionally, we propose a language-guided feature aggregation (LGFA) module that fosters interaction between multi-scale visual representations and linguistic features, thereby capturing cross-scale dependencies and addressing complex scale variations. To further enhance feature aggregation, the LGFA incorporates an attention deficit compensation mechanism. Finally, we design a mutual-interaction decoder (MID) to achieve precise vision-language alignment via cascaded bidirectional cross-attention, ultimately generating highly accurate segmentation masks.

To further advance research in RRSIS, we construct a new dataset called RISBench, with images sourced from the DOTA-v2\cite{ding2021object} and DIOR\cite{li2020object} remote sensing object detection datasets. RISBench consists of 52,472 image-language-label triplets. The language expressions provide not only basic category information but also details on color, shape, location, size, relative position, and relative size, with an average length of 14.31 words. Additionally, we employed a semi-automated approach to generate pixel-level mask annotations using bounding box prompts from the VRSBench dataset\cite{li2024vrsbench} and a customized segment anything model (SAM)\cite{kirillov2023segment}. Compared to the RefSegRS and RISBench datasets, our RISBench offers a greater number of triplets, a wider range of spatial resolutions, and a richer diversity of objects within each category.

In summary, the contributions of this work can be summarized in the following three aspects:

\begin{itemize}
	
	\item[(1)] We propose CroBIM, a novel framework specifically designed to address the unique challenges of referring remote sensing image segmentation (RRSIS), including the diversity of remote sensing data and the complexity of geospatial relationships described in natural language.
	
	\item[(2)] We design three task-specific modules to address the core challenges of RRSIS: the context-aware prompt modulation (CAPM) integrates multi-scale visual context with linguistic features using learnable prompts, the language-guided feature aggregation (LGFA) captures cross-scale dependencies to enhance feature aggregation, and the mutual-interaction decoder (MID) ensures precise vision-language alignment through cascaded bidirectional cross-attention, collectively enabling accurate segmentation mask predictions.
	
	\item[(3)] To advance RRSIS research, we construct RISBench, the largest and most diverse benchmark dataset to date, consisting of 52,472 high-quality image-language-label triplets. RISBench features detailed referring expressions, diverse remote sensing scenarios, and pixel-level masks generated through a robust semi-automatic process, providing a comprehensive resource for model training and evaluation.

	\item[(4)] Existing referring image segmentation methods are extensively evaluated on three benchmark datasets. The experimental results robustly validate the effectiveness and generalization capabilities of our proposed approach, demonstrating its superior performance in comparison to state-of-the-art (SOTA) methods.
	
\end{itemize}

The remainder of this paper is structured as follows: Section~\ref{section:Related Work} reviews related works on RRSIS. Section~\ref{section:Dataset Construction} details the construction process of our proposed RISBench dataset and provides an analysis of its key characteristics. In Section~\ref{section:methods}, we describe the proposed methodology in detail. Section~\ref{section:EXPERIMENTS} presents a comprehensive set of experiments and in-depth analyses. Finally, Section~\ref{section:Conclusion} concludes the paper and offers insights into potential future research directions.

\section{Related Work}
\label{section:Related Work}

\subsection{Referring Image Segmentation}

Compared to other multimodal tasks, referring image segmentation is more challenging as it requires effective coordination and reasoning between language and vision to accurately segment the target regions in an image. Multimodal fusion, diversity of expression, and robustness are three critical challenges that need to be addressed in the current state of referring image segmentation tasks\cite{ji2024survey}.


Hu \textit{et al.}\cite{hu2016segmentation} proposed a CNN-LSTM framework for referring image segmentation, effectively extracting visual and linguistic features for precise segmentation. A recurrent refinement network (RRN)\cite{li2018referring} was proposed to capture multi-scale semantics in image representations. The RRN iteratively optimized the initial mask using a recursive optimization module to achieve a high-quality pixel segmentation mask.


However, these methods often neglect the modality gap and fail to fully capture the interaction between language and images. Recent works have introduced attention mechanisms to address these limitations. A cross-modal self-attention (CMSA) module by Ye \textit{et al.}\cite{ye2019cross} was proposed to effectively captures long-range dependencies between language and visual features. A cascade-grouped attention network (CGAN)\cite{luo2020cascade} is designed, consisting of cascade-grouped attention (CGA) and instance-level attention loss (ILA). By performing hierarchical reasoning on images and effectively distinguishing different instances, CGAN enhances the correlation between text and images. Besides, Hu \textit{et al.}\cite{hu2020bi} introduced a bidirectional relationship inferring network (BRINet) to model cross-modal information dependencies. BRINet utilized a visual-guided language attention module to filter out irrelevant regions and enhance semantic matching between target objects and expressions.

\begin{figure*}[tbp]
	\begin{center}
		\centerline{\includegraphics[width=1\linewidth]{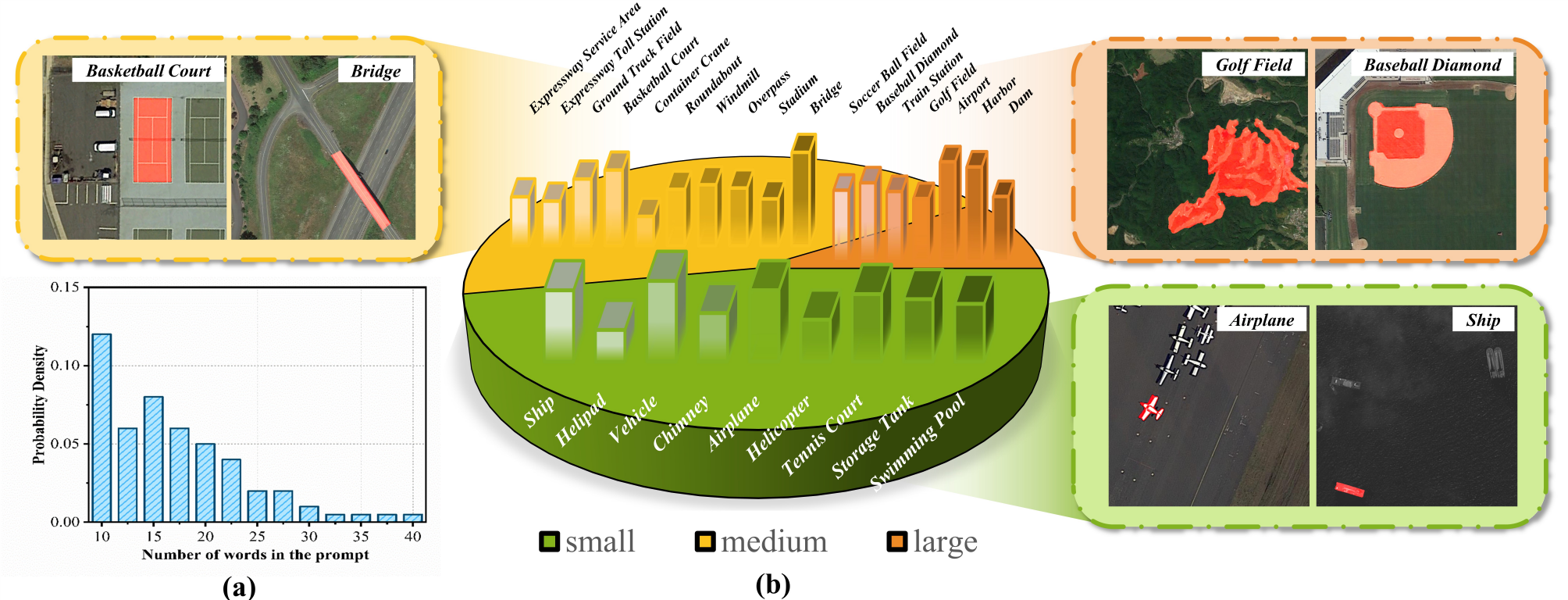}}
		\caption{Statistical analysis of the constructed RISBench dataset. (a) Distribution of the word length of referring expressions. (b) Distribution of the object categories and object size.}\label{statistics}
	\end{center}
\end{figure*}

\subsection{Visual Grounding for Remote Sensing Data}

Similar to RRSIS, visual grounding for remote sensing data (RSVG) specifically entails using a remote sensing image alongside an associated query expression to determine the bounding box for a target object of interest. By localizing objects in remote sensing scenes through natural language guidance, RSVG provides object-level understanding and enhances accessibility. Compared to query expressions in natural images, expressions in RSVG frequently encompass complex geospatial relationships, and the objects of interest are often not visually prominent.

GeoVG\cite{sun2022visual} is the first RVSA framework, which utilizes a language encoder to learn spatial relationships in geographic space, an image encoder to adaptively attend to remote sensing scenes, and a fusion module to integrate textual and visual features for visual localization. Zhan \textit{et al.}\cite{zhan2023rsvg} proposed a large-scale benchmark dataset DIOR-RSVG, and designed a Transformer-based multigranularity visual language fusion (MGVLF) module is proposed, which addresses the challenges of large-scale variations and cluttered backgrounds in remotely sensed images. By leveraging multiscale visual features and multigranularity textual embeddings, more discriminative representations are learned. Besides, language-guided progressive visual attention framework (LPVA)\cite{li2024language} utilized a progressive attention module to adjust visual features at multiple scales and levels, enabling the visual backbone to focus on expression-related features. Additionally, a multi-level feature enhancement decoder aggregated visual contextual information, enhancing feature distinctiveness and suppressing irrelevant regions.

\subsection{Referring Remote Sensing Image Segmentation}

Referring image segmentation in the context of remote sensing data has emerged as a novel area of investigation in recent times. Studies pertaining to this specific task are currently in an ascent stage and remain relatively scarce. Yuan \textit{et al.}\cite{yuan2024rrsis} first introduced the RRSIS task in the remote sensing domain. To facilitate research on RRSIS, they constructed a benchmark dataset RefSegRS by designing various referring expressions and automatically generating corresponding masks. Specifically, the RefSegRS dataset consists of 4,420 image-language-label triplets. Furthermore, to address the challenge of segmenting small and scattered objects in remote sensing images, they devised a language-guided cross-scale enhancement (LGCE) module based on the language-aware vision Transformer (LAVT)\cite{yang2022lavt}. The LGCE module leveraged linguistic features as guidance to improve the segmentation of small objects by integrating deep and shallow features, thereby enhancing the complexity and diversity of the approach. In addition, to address the spatial variations and rotational diversity of targets in aerial images, the rotated multi-scale interaction network (RMSIN)\cite{liu2024rotated} was proposed. RMSIN introduced the intra-scale interaction module (IIM) and cross-scale interaction module (CIM) within the LAVT framework, enabling the extraction of detailed features and facilitating comprehensive feature fusion. Moreover, to effectively handle the intricate rotational variations of objects, the decoder of RMSIN integrated the adaptive rotated convolution (ARC). This integration enhances the network's capability to capture and represent complex object rotations, thereby improving the overall performance on the RRIS task.

\section{Dataset Construction and Analysis}
\label{section:Dataset Construction}

In this section, we will introduce the construction procedure and statistical analysis of our proposed RISBench dataset in Section~\ref{RISBench dataset Construction} and Section~\ref{Dataset Statistics}.

\subsection{RISBench Dataset Construction}
\label{RISBench dataset Construction}

Motivated by the SAM\cite{kirillov2023segment} and RMSIN\cite{liu2024rotated}, we combine bounding box prompts with the SAM to generate pixel-level masks using a semi-automatic method, significantly reducing the cost of manual annotation. The steps for generating fine-grained pixel-level annotations for the RRSIS task are as follows:

\begin{itemize}

	\item \textbf{Step 1.} We collected remote sensing images, referring textual descriptions, and corresponding visual grounding boxes from the VRSBench dataset \cite{li2024vrsbench}. However, the bounding boxes often exhibited inaccuracies, such as misalignments and inappropriate sizing (either too large or too small). Additionally, there exists a significant domain gap between natural and remote sensing scenes, which exacerbates the problem. Consequently, directly applying these bounding boxes to the SAM model results in unsatisfactory segmentation masks, necessitating further refinement and optimization.
	
	\item \textbf{Step 2.} To ensure the accuracy and reliability of the generated mask annotations, we employed the PA-SAM method \cite{xie2024pa} to optimize the SAM model. Specifically, we trained a specialized adapter using a subset of remote sensing images with manually corrected masks, enabling SAM to better handle the unique characteristics of remote sensing imagery, such as complex geospatial relationships and varying object scales. The fine-tuning process involved using bounding box prompts as input, optimizing the mask decoder features with a combination of Dice loss and cross-entropy loss, and iteratively refining the model through multiple rounds of training and validation. After generating the initial masks, we conducted a meticulous human verification process, where expert annotators manually reviewed and corrected the masks. This process included initial inspections, detailed corrections using specialized tools, and a consensus process for uncertain cases, ensuring high-quality and objective annotations.
	


\begin{table}[tbp]
	\centering
	\footnotesize
	\caption{Comparative analysis of RISBench and prior RRSIS datasets.}
	\label{dataset}
	\renewcommand{\arraystretch}{1.4}
	\begin{tabular}{@{}c|c|c|c|c@{}}
		\toprule
		Dataset & \makecell{Triplet \\ Count} & \makecell{Image \\ Size} & \makecell{Spatial \\ Resolution} & \makecell{Expression \\ Attributes} \\
		\midrule
		RefSegRS\cite{yuan2024rrsis}   & 4,420   & 512×512     & 0.13\,m        & 3 \\
		RRSIS-D\cite{liu2024rotated}   & 17,402  & 800×800     & 0.5\,m $\sim$ 30\,m  & 7 \\
		RISBench                      & 52,472  & 512×512     & 0.1\,m $\sim$ 30\,m  & 8 \\
		\bottomrule
	\end{tabular}
\end{table}

	\item \textbf{Step 3.} To enhance the segmentation masks, we employ a meticulous human verification process. Expert annotators manually review and correct the masks generated by the optimized SAM model. Initially, annotators identify inaccuracies or misalignments based on their expertise and predefined criteria. Detailed inspections are then conducted on flagged masks, where annotators assess boundary precision and object shapes by zooming in on specific image regions. Using specialized tools, they refine contours, correct misalignments, resize segments, and resolve boundary ambiguities. When initial corrections are uncertain, a consensus process is initiated, involving independent reviews by multiple annotators and final decisions through majority agreement or expert discussion. This human-in-the-loop approach ensures high-accuracy mask annotations, bridging the gap between automated segmentation and the nuanced understanding required for remote sensing imagery. 
\end{itemize}

While potential biases may exist in the dataset due to factors such as annotator subjectivity, dataset characteristics, and linguistic variations, we have taken several steps to minimize their impact. During the annotation process, we employed a consensus-based verification approach, where multiple expert annotators independently reviewed and corrected the masks, ensuring objectivity and accuracy. Additionally, the fine-tuning of the SAM model and the use of diverse remote sensing images helped reduce biases related to object scales and spatial relationships. These measures ensure that the dataset is robust and representative, enhancing the model's generalization ability to other datasets and real-world scenarios.

\begin{figure}[tbp]
	\begin{center}
		\centerline{\includegraphics[width=1\linewidth]{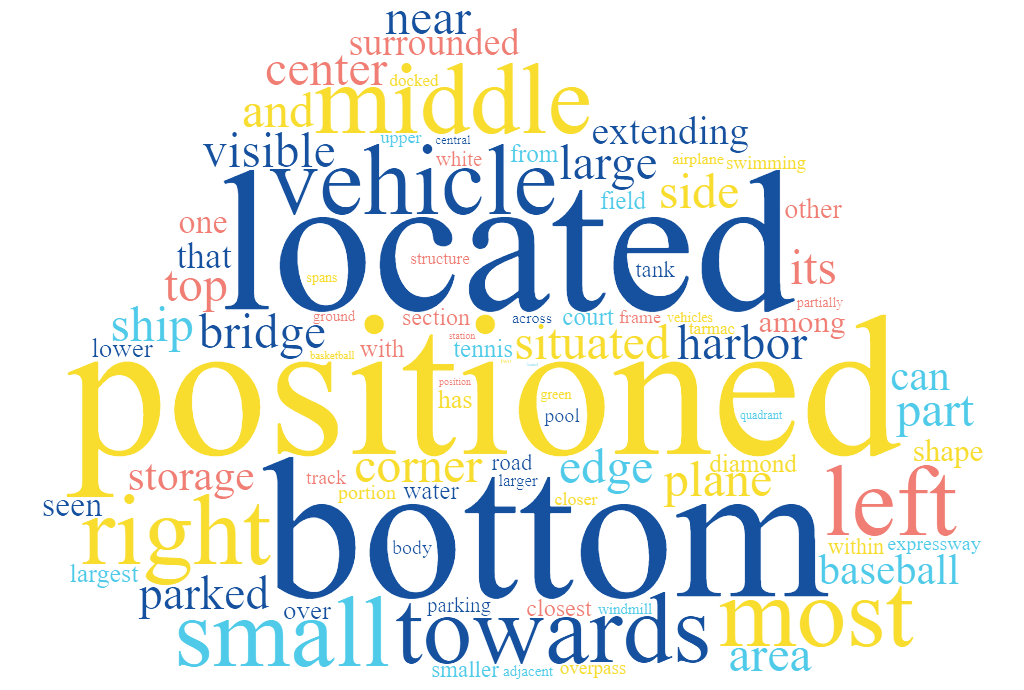}}
		\caption{Word cloud for top 50 words within the referring sentences in our RISBench dataset. }\label{wordcloud}
	\end{center}
\end{figure}

\begin{figure*}[tbp]
	\begin{center}
		\centerline{\includegraphics[width=1\linewidth]{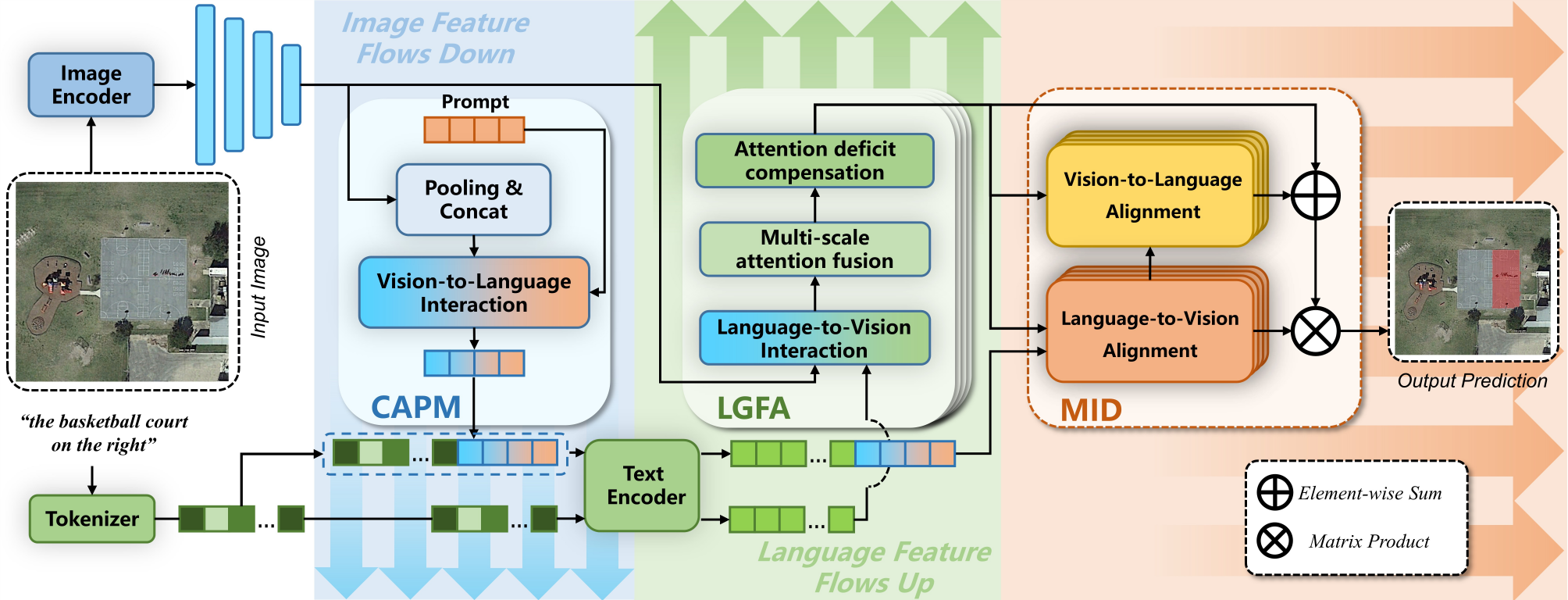}}
		\caption{Overview of our proposed CroBIM framework, which comprises five key components: an image encoder, an text encoder, context-aware prompt modulation (CAPM) module, language-guided feature aggregation (LGFA) module, and mutual-interaction decoder (MID).}\label{flowchart}
	\end{center}
\end{figure*}

\subsection{Dataset Statistics}
\label{Dataset Statistics}

After meticulously filtering out uninformative image-language label triplets, we curated the RISBench dataset, comprising 52,472 high-quality image-language label triplets. This dataset is partitioned into a training set with 26,300 triplets, a validation set with 10,013 triplets and a test set with 16,158 triplets, ensuring robust model development and evaluation. Each image in RISBench is uniformly sized at 512$\times$512 pixels, maintaining consistency across the dataset. The spatial resolution of the images spans from 0.1m to 30m, encompassing a diverse range of scales and details. The semantic labels are categorized into 26 distinct classes, each annotated with 8 attributes, thereby facilitating a comprehensive and nuanced semantic segmentation analysis. As shown in Table.~\ref{dataset}, compared to the previous RRSIS datasets, our dataset demonstrates significant improvements in both quantity and diversity.

Additionally, the distribution of categories and object sizes is illustrated in Fig.~\ref{statistics}(b), respectively. Moreover, the referring expressions in our dataset have an average length of 14.31 words, and the vocabulary size encompasses 4,431 unique words, underscoring the richness and complexity of the language component. The distribution of word lengths within these expressions is depicted in Fig.~\ref{statistics}(a), providing further insight into the linguistic characteristics of the dataset. Fig.~\ref{wordcloud} illustrates the word cloud representation of the RISBench dataset.

\section{Methodology}
\label{section:methods}

The overall architecture of our proposed framework is shown in Fig.~\ref{flowchart}. Our CroBIM framework processes an input image $I\in\mathbb{R}^{H \times W \times 3}$
and a language expression $E=\left\{e_i\right\}, i \in\{0, \ldots, N\}$, where $H$ and $W$ denote the height and width of the input image, respectively, and $N$ represents the length of the referring expression. For the image encoder, we employ one of Swin Transformer\cite{liu2021swin} and ConvNeXt\cite{liu2022convnet}, which extracts multi-scale visual features $\left\{V \in \mathbb{R}^{H_i \times W_i \times C_i}\right\}_{i=1}^4$ from the input image. Here,  $\left(H_i, W_i\right)=\left(H / 2^{i+1}, W / 2^{i+1}\right)$ and $C_i$ denote the spatial resolution and channel dimension of the $i$-th visual feature, respectively. Additionally, We employ BERT \cite{devlin2018bert} as the text encoder to process the referring description by tokenizing and padding it, which generates text tokens $T \in \mathbb{R}^{l_m \times D_l}$. These tokens are then input into the BERT encoder to derive linguistic features $L \in \mathbb{R}^{l_m \times D_l}$. In this context, $l_m$ denotes the maximum token length, while $D_l$ represents the dimension of the linguistic features.

The details of each part will be introduced in the following sections.

\subsection{Context-Aware Prompt Modulation}
\label{section:submethod1}

\begin{figure}[tbp]
	\begin{center}
		\centerline{\includegraphics[width=1\linewidth]{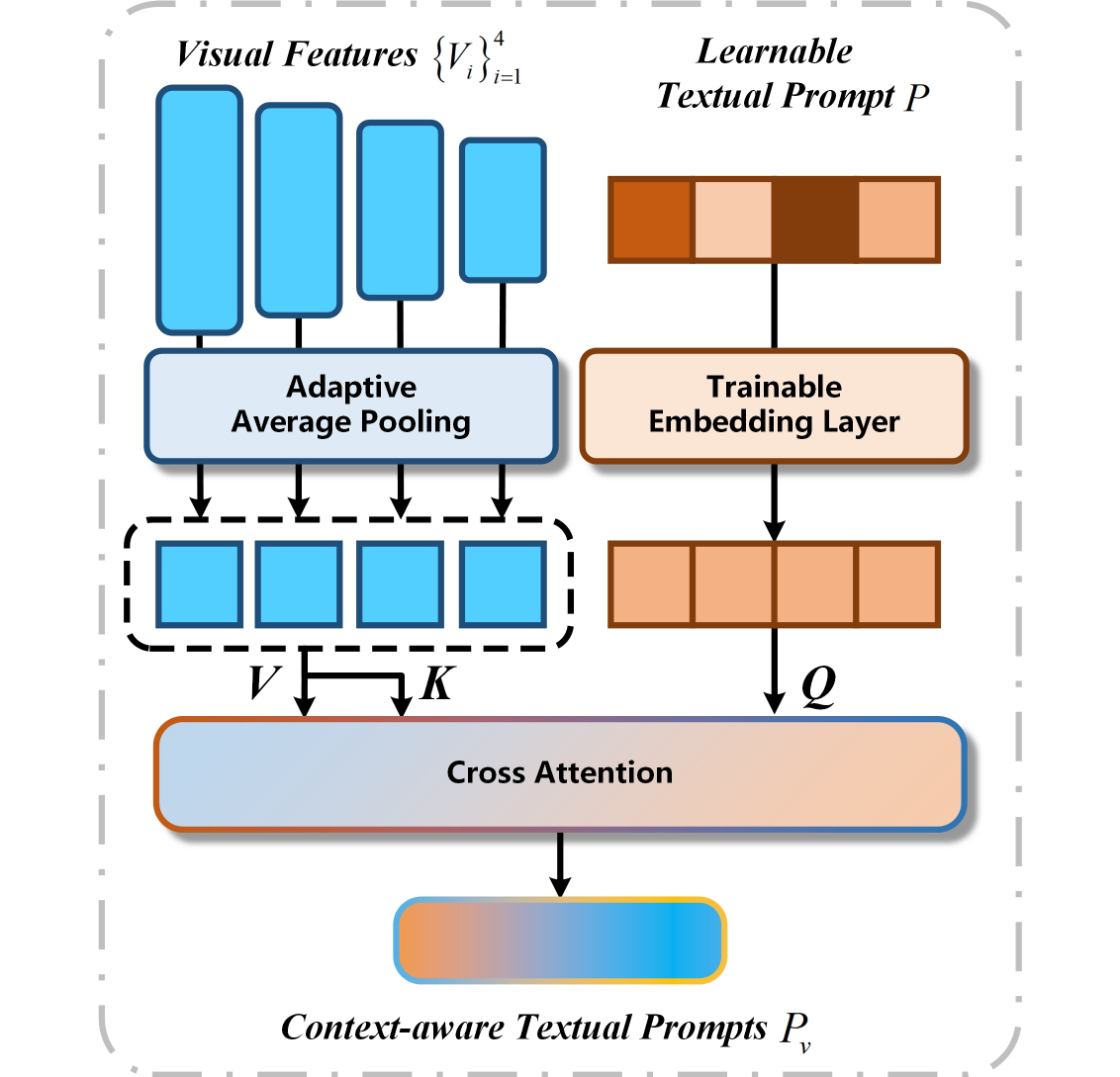}}
		\caption{Pipeline of the context-aware prompt modulation (CAPM) module. By integrating multi-scale visual contextual information through learnable prompts, CAPM enables the model to accurately capture the spatial structure and relative positioning of target objects as described in the referring expressions.}\label{CAPM}
	\end{center}
	\vspace{-5mm} 
\end{figure}

Prompt learning enhances model adaptability to specific tasks by introducing learnable parameters. However, free-form text prompts often lack sufficient contextual information, leading to suboptimal quality of learned representations. To address this issue, we propose a context-aware prompt modulation (CAPM) module, as illustrated in Fig.~\ref{CAPM}. The CAPM module integrates multi-scale visual contextual information during the text encoding process, aiding the model in better perceiving the spatial structure and relative positioning of target objects, thereby improving its ability to capture and identify these objects effectively.

Given the multi-scale visual features $\left\{V_i \right\}_{i=1}^4$ produced by the image encoder, we first apply adaptive average pooling to extract cross-scale contextual information. Subsequently, the pooled features from each scale are concatenated and flattened to form a multi-scale context embedding $V_e$:
\begin{equation}
	V_e=\text {Flatten}\left(\text {Concat}\left(\left\{\operatorname{Pool}_{\mathrm{s} \times \mathrm{s}}\left(V_i\right)\right\}_{i=1}^4\right)\right) \in \mathbb{R}^{4 s^2 \times C_{total}},
\end{equation}where $\operatorname{Concat(\cdot)}$ denotes the channel concatenation operation, $\text{Flatten}(\cdot)$ converts a multidimensional tensor into a one-dimensional vector, and $\operatorname{Pool_{s \times s}}$ represents adaptive average pooling with an output size of $s \times s$. Here, $C_{total}=\sum_{i=1}^4 C_i$, and $s$ is set to 1 in this work.

Furthermore, learnable textual prompts $P \in \mathbb{R}^{N_p \times D_l}$ are introduced as supplementary inputs to guide the model in incorporating domain-specific knowledge pertinent to RRSIS task into the learning process, where $N_p$ is set to 4. This enhancement aims to improve the model's capability to comprehend and generate responses relevant to the current task. To achieve image-to-text cross-modal interaction, we introduce cross attention mechanism to integrate multi-scale context into the learnable textual prompts $P$ with $V_e$:
\begin{equation}
	P_v
	=\operatorname{CrossAttn}(V_e, P) \\
	=\operatorname{Softmax}\left(P \omega_q\left(V_e \omega_k\right)^{\top}\right) V_e \omega_v,
\end{equation}where $\omega_q,\omega_k,\omega_v$ are the projection matrices, and $P_v$ represents the context-aware textual prompts.

The context-aware textual prompts $P_v$ are then concatenated with the text tokens $T$ and jointly input into the BERT encoder to obtain linguistic features $L_v \in \mathbb{R}^{\left(l_m+N_p\right) \times D_l}$ that incorporate visual context.

\subsection{Language-Guided Feature Aggregation}

\begin{figure}[tbp]
	\begin{center}
		\centerline{\includegraphics[width=1\linewidth]{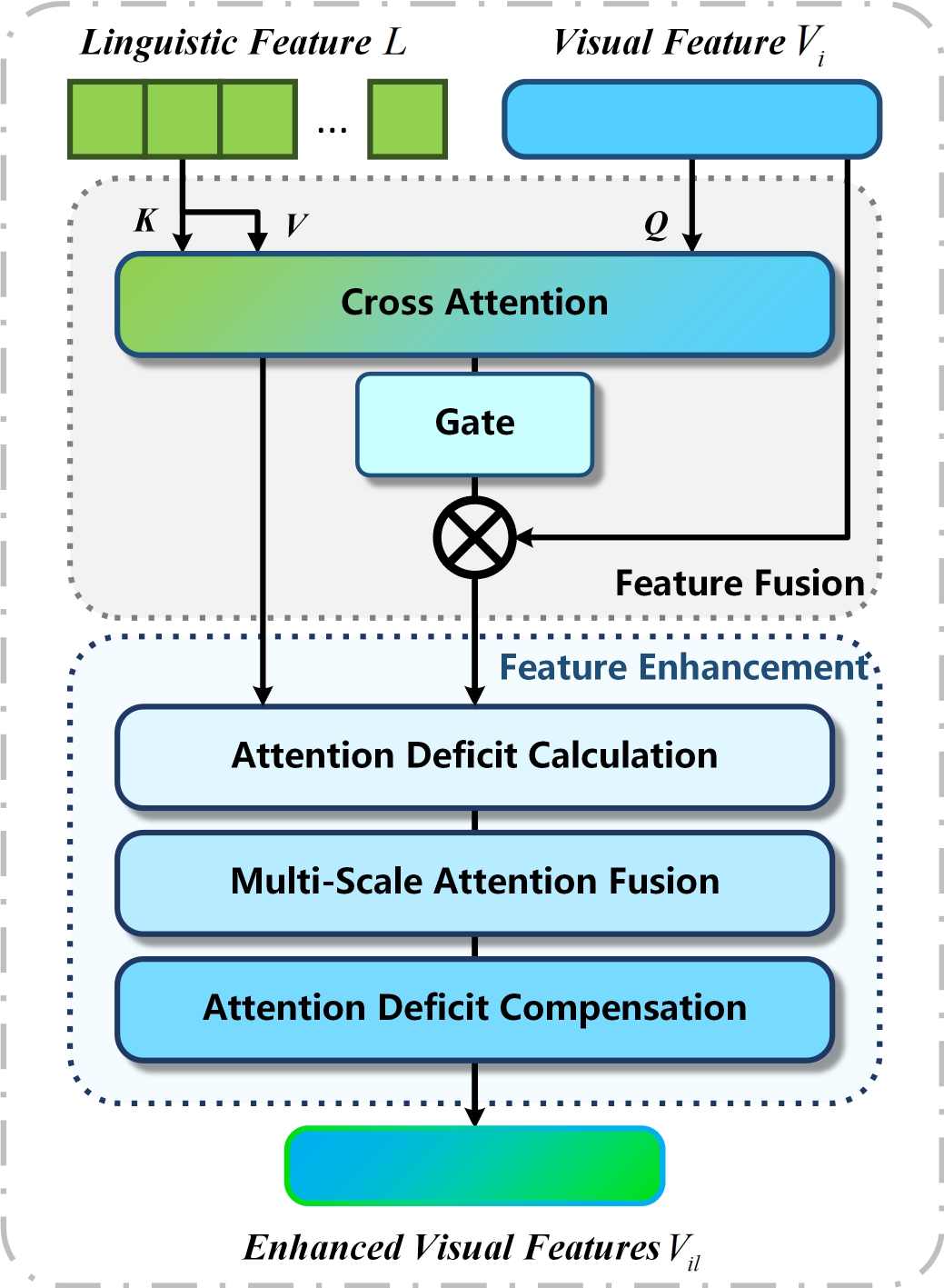}}
		\caption{An illustration of our proposed language-guided feature aggregation (LGFA) module. After performing cross-modal feature fusion, attention deficit compensation is introduced to refine multi-scale visual features using textual constraints, thereby ensuring that attention across different stages is focused on the same target region.}\label{LGFA}
	\end{center}
\end{figure}

To forge a robust visual-linguistic synergy and seamlessly integrate dependable linguistic features of referred objects into multi-scale visual representations, we introduce a sophisticated language-guided feature aggregation (LGFA) module. As shown in Fig.~\ref{LGFA}, the LGFA module adeptly captures and models the intricate interdependencies between visual and linguistic modalities.

Initially, the visual feature $V_i \in \mathbb{R}^{H_i \times W_i \times C_i}$ undergoes processing through a projection function $\omega_{i q}$. Subsequently, it is spatially expanded, as delineated by:
\begin{equation}
	V_{iq} = \mathrm{Flatten}(\omega_{iq}(V_i)),
\end{equation}where $\omega_{i q}$ denotes the projection function for the visual feature and $V_{i q} \in \mathbb{R}^{l_m \times (H_i \times W_i)}$ represents the projected and spatially expanded visual feature.

Subsequently, the linguistic feature $L \in \mathbb{R}^{l_m \times D_l}$  undergoes transformation through projection functions $\omega_{ik}$ and $\omega_{iv}$:
\begin{equation}
	L_{ik}, L_{iv} = \omega_{ik}(L), \omega_{iv}(L),
\end{equation}where $\omega_{ik}$ and $\omega_{iv}$ function as projection mechanisms for the linguistic features, with $L_{ik}$ and $L_{iv}$ representing the linguistic features utilized for computing the attention scores and generating the final attention output, respectively.

Following this, we compute the attention score matrix $S_i$ between the visual and linguistic features:
\begin{equation}
	S_i = V_{iq}^\mathrm{T} L_{ik} \in \mathbb{R}^{(H_i \times W_i) \times D_l},
\end{equation}

Afterward, the attention score matrix $S_i$ is normalized using the Softmax function, multiplied by $L_{iv}^T$,, and finally, gated cross-modal activation is obtained by applying a Gate operation subsequent to the unflatten operation:
\begin{equation}
	\operatorname{Att}_i = \mathrm{Gate}(\mathrm{Unflatten}(\operatorname{Softmax}\left(\frac{S_i}{\sqrt{l_m}}\right) L_{iv}^T)),
\end{equation}where $\mathrm{Gate}(\cdot) $ represents the application of a $1 \times 1$ convolution followed by a GELU activation function, while $\mathrm{Unflatten}(\cdot)$ indicates the inverse operation of $\mathrm{Flatten}(\cdot)$.

Finally, the input visual feature $V_i$ is reweighted by integrating the attention weights $\operatorname{Att}_i$, resulting in the integrated cross-modal feature map $F_i \in \mathbb{R}^{H_i \times W_i \times C_i}$:
\begin{equation}
	V_{li}=\operatorname{Conv}_{1 \times 1}\left(A t t_i\right) \odot V_i,
\end{equation}where $\odot$ denotes element-wise matrix multiplication, and $\operatorname{Conv}_{1 \times 1}$ represents the $1\times1$ convolution function.

To align the cross-modal correlations between adjacent stages and refine the aggregated multi-scale features, we resample the attention maps $\left\{S_i \right\}_{i=1}^4$ of different scales to a uniform size $(H_4, W_4)$:
\begin{equation}
	S_i' = \mathcal{I}(S_i, (H_4, W_4)),i \in \{1, 2, 3, 4\},
\end{equation}where $\mathcal{I}$ denotes the resampling operation.

Subsequently, we introduce an attention deficit compensation mechanism that identifies regions where attention diverges across different scales and explicitly enhances these areas by integrating multi-scale visual features guided by linguistic cues. Specifically, we calculate the attention deficit map $\mathcal{M}\in \mathbb{R}^{H_4 \times W_4}$ between cross-scale attentions and select the top $K$ regions with the highest correlation differences:
\begin{equation}
	\mathcal{M} = \sum_{i=1}^{3} \left| s_i - s_{i+1} \right|,
\end{equation}
\begin{equation}
	\{ \mathbf{r}_{k}\}_{k=1}^{K} \leftarrow
	\text{TopK}\left(\mathcal{M}, K\right),
\end{equation}

For each attention deficit region $\mathbf{r}_{k}$, the multi-scale features $\left\{V_{li}^{r_k}\right\}_{i=1}^4$ corresponding to that region are projected to a uniform channel dimension $C_{\hat v}$ and concatenated to yield the cross-scale feature representation $F_{\text {cs}}^{\mathbf{r}_k}$:
\begin{equation}
	V_{l}^{\mathbf{r}_k}=\operatorname{Concat}\left(\overrightarrow{\operatorname{Proj}}\left(\mathcal{I}\left[V_{l1}^{\mathbf{r}_k}, V_{l2}^{\mathbf{r}_k}, V_{l3}^{\mathbf{r}_k}, V_{l4}^{\mathbf{r}_k}\right]\right)\right),
\end{equation}where $\operatorname{Concat}(\cdot)$ denotes the concatenation operaion, and $\overrightarrow{\operatorname{Proj}}(\cdot)$ indicates the channel projection layer. 

We then proceed to characterize cross-scale dependencies with the following steps:
\begin{equation}
	\widetilde{V}_{l}^{\mathbf{r}_k}=\operatorname{MSA}\left(\operatorname{LN}\left(V_{l}^{r_k}\right)\right)+V_{l}^{r_k},
\end{equation}where $\operatorname{LN}(\cdot)$ denotes the layer normalization operator, and $\operatorname{MSA}(\cdot)$ signifies the multi-head self-attention mechanism. 

Following this, the enhanced sequence is meticulously restructured back into its original patch configuration, precisely adhering to the initial order of concatenation, ensuring a seamless and coherent transformation:
\begin{equation}
	\left[\widetilde{V}_{l1}^{\mathbf{r}_k}, \widetilde{V}_{l2}^{\mathbf{r}_k}, \widetilde{V}_{l3}^{\mathbf{r}_k}, \widetilde{V}_{l4}^{\mathbf{r}_k}\right]=\mathcal{I}^{\prime}\left(\operatorname{Split}\left(\overleftarrow{\operatorname{Proj}}(\widetilde{V}_{l}^{\mathbf{r}_k})\right)\right),
\end{equation}where $\overleftarrow{\operatorname{Proj}}(\cdot)$ and $\mathcal{I}^{\prime}$ represents the inverse operation of $\overrightarrow{\operatorname{Proj}}(\cdot)$ and $\mathcal{I}$, and $\operatorname{Split}(\cdot)$ represents the channel separation operation.

\subsection{Mutual-Interaction Decoder}

\begin{figure}[tbp]
	\begin{center}
		\centerline{\includegraphics[width=1\linewidth]{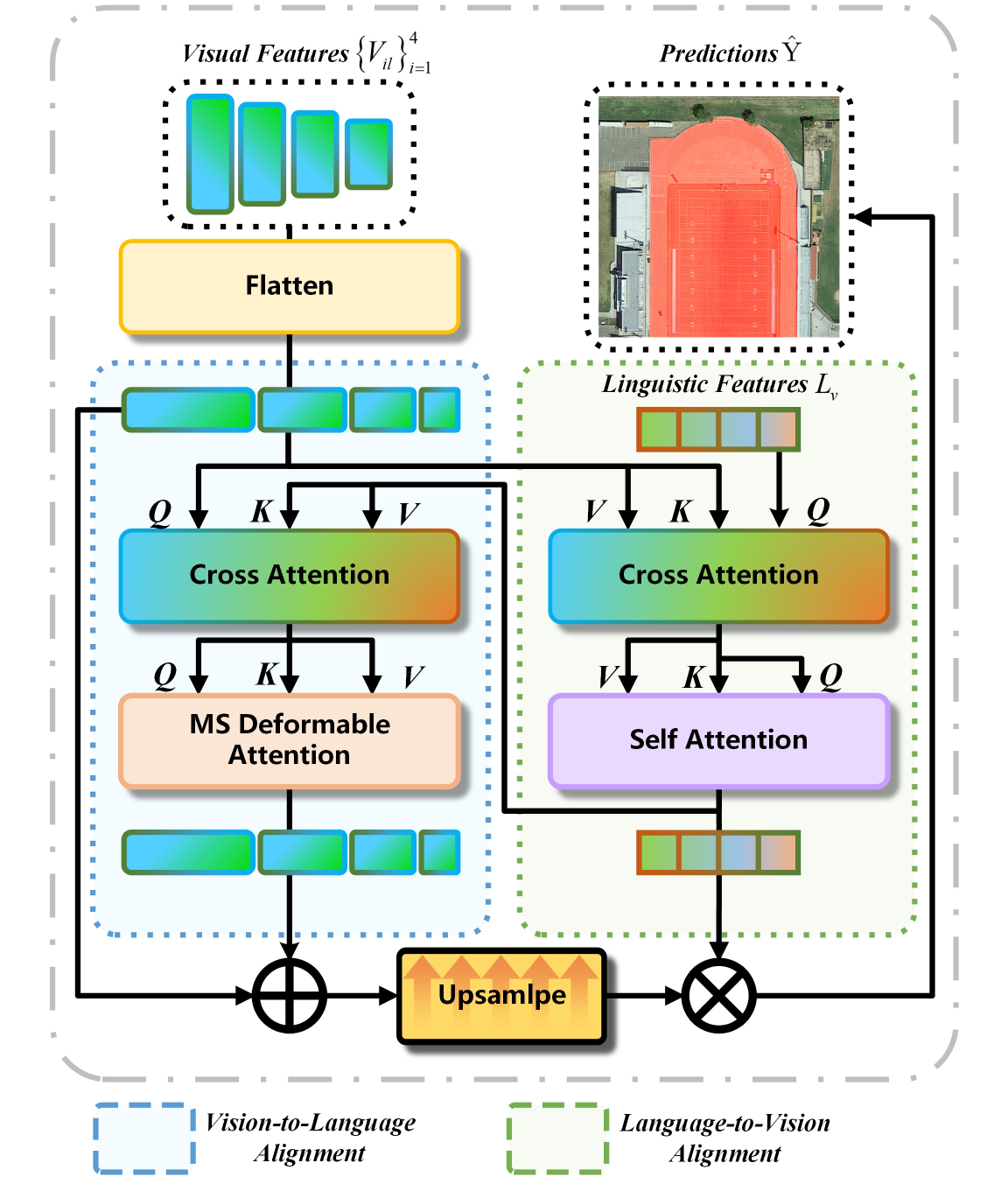}}
		\caption{Illustration of the mutual-interaction decoder (MID) , which incorporates visual context into language features via vision-to-language alignment, followed by aligning these enriched language features with individual visual pixels through language-to-vision alignment. The aligned visual-linguistic features are then used to accurately segment the target objects.}\label{MID}
	\end{center}
\end{figure}

Integrating the complementary information between cross-modal features is a fundamental challenge in RRSIS task. Cross-modal features often encompass inconsistent information, and without considering the intermediate interactions and thorough alignment between different modalities, it is impossible to ensure the discriminative power of the learned representations. To address the aforementioned challenges, we have meticulously designed the mutual-interaction decoder (MID), as illustrated in Fig.~\ref{MID}, aiming to achieve more comprehensive vision-language alignment and precise pixel prediction.

The MID utilizes visual features $\left\{V_{il}\right\}_{i=1}^4$ and linguistic features $L_v$ as inputs to predict the mask of the referred object. Prior to executing vision-language alignment, a series of operations are performed to harmonize the dimensions of the visual and linguistic features.
\begin{equation}
	V_{ms}=\operatorname{Flatten}\left(\operatorname{Concat}\left(\operatorname{Proj}_v\left(\left\{V_{il}\right\}_{i=1}^4\right)\right)\right) \in \mathbb{R}^{N \times D},
\end{equation}where $\operatorname{Proj}_v$ projects the multi-scale visual features into a hidden dimension $D=256$, and $N=\sum_{i=1}^4 H_i W_i$.

Specifically, given the linguistic features $L_v$ and visual features $V_{ms}$, the language-to-vision interaction is facilitated through cross-attention, self-attention, and feed-forward networks (FFN). These mechanisms are employed to update the linguistic features $L_v$ using $V_{ms}$. Each layer of cross-attention, self-attention, and FFN is followed by a residual connection and layer normalization, ensuring a coherent and stable transformation of the features.
\begin{equation}
	\hat{L}_{v}=\operatorname{FFN}(\operatorname{SelfAttn}(\operatorname{CrossAttn}(L_v, V_{ms}))),
\end{equation}

Subsequently, the refined linguistic features $\hat{L}_{v}$ are meticulously aligned with the visual features $V_{ms}$ on a pixel-by-pixel basis through a sophisticated vision-to-language interaction mechanism:
\begin{equation}
	\hat{V}_{ms}=\operatorname{FFN}(\operatorname{MSDeformAttn}(\operatorname{CrossAttn}(V_{ms}, \hat{L}_v))),
\end{equation}where $\operatorname{MSDeformAttn}(\cdot)$ denotes the multiscale deformable attention\cite{zhu2020deformable}.

Upon executing mutual interactions through bidirectional cross-modal feature alignment, we derive the harmonized linguistic and visual features, denoted as $\hat{L}_{v}$ and $\hat{V}_{ms}$.

Finally, the enhanced visual features $\hat{V}_{ms}$ are combined with the original visual features $V_{ms}$ and subsequently mapped through through a 1$\times$1 convolutional layer followed by spatial resampling to obtain the mask embedding $V_{\text{out}} \in \mathbb{R}^{H_1 \times W_1 \times D}$. $V_{\text{out}}$ is then element-wise multiplied with the [CLS] token $L_{\text{out}} \in \mathbb{R}^{D}$ of the linguistic features $\hat{L}_{v}$, resulting in the final prediction mask $\operatorname{\hat{Y}} \in \mathbb{R}^{H_1 \times W_1}$.
\begin{equation}
	\operatorname{\hat{Y}}^{(i, j)}=V_{\text {out}}^{(i, j)} \cdot L_{\text {out}}.
\end{equation}where $(i, j)$ represents the pixel position in a two-dimensional space. $\operatorname{\hat{Y}}$ is then upsampled to the same spatial resolution $(H, W)$ as the input image via bilinear interpolation.

Unlike existing methods, which typically rely on unidirectional or static interactions that may lead to attention drift or misalignment between modalities, our proposed MID introduces a bidirectional refinement process. This process enables continuous and iterative alignment between visual and linguistic modalities throughout the decoding stage. Such bidirectional interaction ensures that both modalities are dynamically optimized, resulting in more precise feature integration and robust segmentation performance, particularly in complex remote sensing scenarios characterized by diverse geospatial relationships.

\subsection{Training Objective}

In the RRSIS task, object mask prediction is typically framed as a pixel-wise binary classification problem. Due to the significant class imbalance in remote sensing images, where target pixels are relatively scarce compared to background pixels, a conventional cross-entropy loss function may lead to a model that prioritizes learning from background pixels, adversely affecting the performance in detecting target regions. To address this issue, we employ a combined loss function consisting of cross-entropy loss and dice loss\cite{li2019dice} as our training objective:

\begin{equation}
	\mathcal{L}=\lambda \cdot \mathcal{L}_{\text {cross-entropy}}(\operatorname{\hat{Y}_{up}}, \operatorname{Y})+ (1-\lambda) \cdot \mathcal{L}_{\text {dice}}(\operatorname{\hat{Y}_{up}}, \operatorname{Y})
\end{equation}
where $\lambda$ is a hyperparameter to balance two losses functions and is set to 0.9 in the paper, $\operatorname{\hat{Y}_{up}}, \operatorname{Y} \in \mathbb{R}^{H \times W}$ represent the upsampled prediction $\operatorname{\hat{Y}}$ and the ground truth, respectively.

\section{Experiments}
\label{section:EXPERIMENTS}

In this section, we perform comprehensive experiments to assess the efficiency and effectiveness of our proposed RRSIS framework.

\subsection{Dataset and Evaluation Metrics}

We conduct experiments on three datasets, including two publicly available datasets and our constructed RISBench dataset. The detailed information for these three datasets is provided as follows:

\textit{\textbf{RefSegRS}}\cite{yuan2024rrsis} comprises 4,420 image-language-label triplets across 285 scenes. The dataset is divided into a training set with 151 scenes and 2,172 referring expressions, a validation set with 31 scenes and 431 expressions, and a test set with 103 scenes and 1,817 expressions. The images are sized at 512$\times$512 pixels, with a spatial resolution of 0.13 meters.

\textit{\textbf{RRSIS-D}}\cite{liu2024rotated} contains a diverse dataset of 17,402 images, each paired with corresponding masks and referring expressions. The dataset is organized into three subsets: a training set with 12,181 image-language-label triplets, a validation set containing 1,740 triplets, and a test set with 3,481 triplets. All images are standardized to a resolution of 800$\times$800 pixels. Additionally, the semantic annotations cover 20 categories and include 7 attributes, thereby enriching the semantic context of the referring expressions.

\textit{\textbf{RISBench}} includes a total of 52,472 image-language-label triplets. It is divided into three subsets: a training set consisting of 26,300 triplets, a validation set consisting of 10,013 triplets and a test set comprising 16,159 triplets. All images are uniformly formatted to a resolution of 512$\times$512 pixels, with spatial resolutions ranging from 0.1 meters to 30 meters. The dataset's semantic labels are divided into 26 unique classes, with each class further annotated by 8 attributes.

Following the prior study\cite{wu2020phrasecut}, overall Intersection-over-Union (oIoU), mean Intersection-over-Union (mIoU), and precision at different threshold values X $\in$ \{0.5,0.6,0.7,0.8,0.9\} (Pr@X) are selected as evaluation metrics.

oIoU is computed as the ratio of the cumulative intersection area to the cumulative union area across all test samples, thereby giving greater emphasis to larger objects:

\begin{equation}
	\text { oIoU }=\left(\sum_t I_t\right) /\left(\sum_t U_t\right)
\end{equation}

In contrast, mIoU is calculated by averaging the IoU values between predicted masks and ground truth annotations for each test sample, treating small and large objects equally:

\begin{equation}
	\text { mIoU }=\frac{1}{M} \sum_t I_t / U_t
\end{equation}where $t$ denotes the index of the image-language-label triplets and $M$ indicates the total size of the dataset. $I_t$ and $U_t$ represent the intersection and union area between the predicted and ground-truth regions.

\subsection{Experimental Setup}

We employ Swin Transformer \cite{liu2021swin} and ConvNeXt \cite{liu2022convnet} as the visual backbones in our approach. The Swin Transformer backbone is initialized with classification weights from the Swin-Base model pre-trained on ImageNet22K \cite{deng2009imagenet}. The ConvNeXt backbone, specifically the ConvNeXt-Base and ConvNeXt-Large variants, is pretrained on the GeoSense dataset, which comprises approximately 9 million diverse remote sensing images, using our proposed SMLFR self-supervised learning framework\cite{dong2024generative}. This pretraining on GeoSense enables the ConvNeXt backbone to learn robust and generalizable feature representations tailored for remote sensing image interpretation. For the language backbone, we use the base BERT model with 12 layers and a hidden size of 768, as available in the HuggingFace library \cite{wolf2020transformers}. The maximum sequence length for text descriptions is set to 20 tokens.

Our method is implemented using PyTorch framework, and we employ the AdamW optimizer \cite{loshchilov2017decoupled} with a weight decay of 0.01 and an initial learning rate of 0.00005. The learning rate is decayed polynomially throughout training. We use a batch size of 32, and each model is trained for 40 epochs on eight NVIDIA A800 GPUs. During both training and testing phases, images are resized to 480$\times$480 pixels. No data augmentation or post-processing techniques are applied.

\begin{table*}[htbp]
	\centering
	\caption{Comparison with state-of-the-art methods on the RRSIS-D dataset. Optimal and sub-optimal performance in each metric are marked by \textcolor{red}{\textbf{red}} and \textcolor{blue}{\textbf{blue}}.}
	\label{rrsisd_comparison}
	\renewcommand\arraystretch{1.4}
	\fontsize{8}{12}\selectfont
	\resizebox{\textwidth}{!}{
		\begin{tabular}{l|c|c|c|c|c|c|c|c|c|c|c|c|c|c|c|c}
			\cmidrule{1-17} 
			\multirow{2}{*}{Method} & \multirow{2}{*}{Visual Encoder} & \multirow{2}{*}{Text Encoder} & \multicolumn{2}{c|}{Pr@0.5} & \multicolumn{2}{c|}{Pr@0.6} & \multicolumn{2}{c|}{Pr@0.7} & \multicolumn{2}{c|}{Pr@0.8} & \multicolumn{2}{c|}{Pr@0.9} & \multicolumn{2}{c|}{oIoU} & \multicolumn{2}{c}{mIoU} \\ \cline{4-17}  
			& & & Val & Test & Val & Test & Val & Test & Val & Test & Val & Test & Val & Test & Val & Test \\ 
			\cmidrule{1-17} 
			RRN\cite{li2018referring} & ResNet-101 & LSTM & 51.09 & 51.07 & 42.47 & 42.11 & 33.04 & 32.77 & 20.80 & 21.57 & 6.14 & 6.37 & 66.53 & 66.43 & 46.06 & 45.64 \\
			CSMA\cite{ye2019cross} & ResNet-101 & None & 55.68 & 55.32 & 48.04 & 46.45 & 38.27 & 37.43 & 26.55 & 25.39 & 9.02 & 8.15 & 69.68 & 69.43 & 48.85 & 48.54 \\
			LSCM\cite{hui2020linguistic} & ResNet-101 & LSTM & 57.12 & 56.02 & 48.04 & 46.25 & 37.87 & 37.70 & 26.35 & 25.28 & 7.93 & 7.86 & 69.28 & 69.10 & 50.36 & 49.92 \\
			CMPC\cite{huang2020referring} & ResNet-101 & LSTM & 57.93 & 55.83 & 48.85 & 47.40 & 36.94 & 35.28 & 25.25 & 25.45 & 9.31 & 9.20 & 70.15 & 69.41 & 51.01 & 49.24 \\
			BRINet\cite{hu2020bi} & ResNet-101 & LSTM & 58.79 & 56.90 & 49.54 & 48.77 & 39.65 & 38.61 & 28.21 & 27.03 & 9.19 & 8.93 & 70.73 & 69.68 & 51.41 & 49.45 \\
			CMPC+\cite{liu2021cross} & ResNet-101 & LSTM & 59.19 & 57.95 & 49.41 & 48.31 & 38.67 & 37.61 & 25.91 & 24.33 & 8.16 & 7.94 & 70.80 & 70.13 & 51.63 & 50.12 \\	
			BKINet\cite{ding2023bilateral} & ResNet-101 & CLIP & 58.79 &	56.9 & 49.54 & 48.77 & 39.65 & 39.12 & 28.21 & 27.03 & 9.19 & 9.16 & 70.78 & 69.89 & 51.14 & 49.65 \\
			ETRIS\cite{xu2023bridging} & ResNet-101 & CLIP & 62.10	& 61.07	& 53.73	& 50.99	& 43.12	& 40.94	& 30.79	& 29.30 & 12.90 & 11.43 & 72.75 & 71.06 & 55.21 & 54.21 \\
			CRIS\cite{wang2022cris} & ResNet-101 & CLIP & 56.44	& 54.84	& 47.87	& 46.77	& 39.77	& 38.06	& 29.31	& 28.15	& 11.84	& 11.52	& 70.98	& 70.46	& 50.75 & 49.69 \\
			LGCE\cite{yuan2024rrsis} & Swin-B & BERT & 68.10 & 67.65 & 60.61 & 61.53 & 51.45 & 51.42 & 42.34 & 39.62 & 23.85 & 22.94 & 76.68 & 76.33 & 60.16 & 59.37 \\
			LAVT\cite{yang2022lavt} & Swin-B & BERT & 65.23 & 63.98 & 58.79 & 57.57 & 50.29 & 49.30 & 40.11 & 38.06 & 23.05 & 22.29 & 76.27 & 76.16 & 57.72 & 56.82 \\
			RMSIN\cite{liu2024rotated} & Swin-B & BERT & 68.39 & 67.16 & 61.72 & 60.36 & 52.24 & 50.16 & 41.44 & 38.72 & 23.16 & 22.81 & \textcolor{red}{\textbf{77.53}} & 75.79 & 60.23& 58.79 \\
			CrossVLT\cite{cho2023cross} & Swin-B & BERT & 67.07 & 66.42 & 59.54 & 59.41 & 50.80 & 49.76 & 40.57 & 38.67 & 23.51 & 23.30 & 76.25 & 75.48 & 59.78 & 58.48 \\
			RIS-DMMI\cite{hu2023beyond} & Swin-B & BERT & 70.40 & 68.74	& 63.05	& 60.96	& \textcolor{blue}{\textbf{54.14}}	& 50.33	& 41.95	& 38.38	& 23.85 & 21.63	& 77.01	& 76.20 & 61.70	& 60.25 \\
			robust-ref-seg\cite{wu2024towards} & Swin-B & BERT & 64.22 & 66.59 & 58.72 & 59.58 & 50.00 & 49.93 & 35.78 & 38.72 & \textcolor{blue}{\textbf{24.31}} & 23.30 & 76.39 & \textcolor{red}{\textbf{77.40}} & 58.92 & 58.91 \\
			MAFN\cite{shi2025multimodal} & Swin-B & BERT & 71.58 & 71.41 & 64.08 & 63.17 & 53.29 & 52.47 & 41.91 & 39.64 & 23.26 & 22.41 & 76.06 & 76.28 & 62.49 & 61.71 \\
			CARIS\cite{liu2023caris} & Swin-B & BERT & 71.61 & 71.50 & 64.66 & 63.52 & \textcolor{blue}{\textbf{54.14}} & 52.92 & \textcolor{blue}{\textbf{42.76}} & 40.94 & 23.79 & \textcolor{red}{\textbf{23.90}} & \textcolor{blue}{\textbf{77.48}} & \textcolor{blue}{\textbf{77.17}} & 62.88 & 62.12 \\ \hline
			CroBIM (Ours) & Swin-B & BERT & \textcolor{blue}{\textbf{74.20}} & \textcolor{red}{\textbf{75.00}} & \textcolor{blue}{\textbf{66.15}} & \textcolor{blue}{\textbf{66.32}} & 54.08 & \textcolor{blue}{\textbf{54.31}} & 41.38 & \textcolor{blue}{\textbf{41.09}} & 22.30 & 21.78 & 76.24 & 76.37 & \textcolor{blue}{\textbf{63.99}} & \textcolor{blue}{\textbf{64.24}}\\  
			CroBIM (Ours) & ConvNeXt-B & BERT & \textcolor{red}{\textbf{74.94}} & \textcolor{blue}{\textbf{74.58}} & \textcolor{red}{\textbf{67.64}} & \textcolor{red}{\textbf{67.57}} & \textcolor{red}{\textbf{57.18}} & \textcolor{red}{\textbf{55.59}} & \textcolor{red}{\textbf{44.66}} & \textcolor{red}{\textbf{41.63}} & \textcolor{red}{\textbf{24.60}} & \textcolor{blue}{\textbf{23.56}} & 76.94 & 75.99	& \textcolor{red}{\textbf{65.05}}	& \textcolor{red}{\textbf{64.46}} \\
			
			\hline
		\end{tabular}
	}
\end{table*}

\begin{table*}[htbp]
	\centering
	\caption{Comparison with state-of-the-art methods on the RefSegRS dataset. Optimal and sub-optimal performance in each metric are marked by \textcolor{red}{\textbf{red}} and \textcolor{blue}{\textbf{blue}}.}
	\label{refsegrs_comparison}
	\renewcommand\arraystretch{1.5}
	\fontsize{8}{12}\selectfont
	\resizebox{\textwidth}{!}{
		\begin{tabular}{l|c|c|c|c|c|c|c|c|c|c|c|c|c|c|c|c}
			\cmidrule{1-17} 
			\multirow{2}{*}{Method} & \multirow{2}{*}{Visual Encoder} & \multirow{2}{*}{Text Encoder} & \multicolumn{2}{c|}{Pr@0.5} & \multicolumn{2}{c|}{Pr@0.6} & \multicolumn{2}{c|}{Pr@0.7} & \multicolumn{2}{c|}{Pr@0.8} & \multicolumn{2}{c|}{Pr@0.9} & \multicolumn{2}{c|}{oIoU} & \multicolumn{2}{c}{mIoU} \\ \cline{4-17}  
			& & & Val & Test & Val & Test & Val & Test & Val & Test & Val & Test & Val & Test & Val & Test \\ 
			\cmidrule{1-17} 
			RRN\cite{li2018referring} & ResNet-101 & LSTM & 55.43 & 30.26 & 42.98 & 23.01 & 23.11 & 14.87 & 13.72 & 7.17 & 2.64 & 0.98 & 69.24 & 65.06 &  50.81 & 41.88 \\
			CMSA\cite{ye2019cross} & ResNet-101 & None & 39.24 & 26.14 & 38.44 & 18.52 & 20.39  & 10.66 & 11.79 & 4.71 & 1.52 & 0.69 & 63.84 & 62.11 & 43.62 & 38.72 \\
			LSCM\cite{hui2020linguistic} & ResNet-101 & LSTM  & 56.82 & 31.54 & 41.24 & 20.41 & 21.85 & 9.51 & 12.11 & 5.29 & 2.51 & 0.84 & 62.82 & 61.27 & 40.59 & 35.54 \\
			BRINet\cite{hu2020bi} & ResNet-101 & LSTM & 36.86 & 20.72 & 35.53 & 14.26 & 19.93 & 9.87 & 10.66 & 2.98 & 2.84 & 1.14 & 61.59 & 58.22 & 38.73  & 31.51 \\
			MAttNet\cite{yu2018mattnet} & ResNet-101 & LSTM & 48.56 & 28.79 & 40.26 & 22.51 & 20.59 & 11.32 & 12.98 & 3.62 & 2.02 & 0.79 & 66.84 & 64.28 & 41.73 & 33.42 \\
			BKINet\cite{ding2023bilateral} & ResNet-101 & CLIP & 52.04 & 36.12 & 35.31 & 20.62 & 18.35 & 15.22 & 12.78 & 6.26 & 1.23 & 1.33 & 75.37 & 63.37 & 56.12 & 40.41 \\
			ETRIS\cite{xu2023bridging} & ResNet-101 & CLIP & 54.99 & 35.77 & 35.03 & 23.00 & 25.06 & 13.98 & 12.53 & 6.44 & 1.62 & 1.10 & 72.89 & 65.96 & 54.03 & 43.11 \\
			CRIS\cite{wang2022cris} & ResNet-101 & CLIP & 53.13 & 35.77 & 36.19 & 24.11 & 24.36 & 14.36 & 11.83 & 6.38 & 2.55 & 1.21 & 72.14 & 65.87 & 53.74 & 43.26 \\
			RMSIN\cite{liu2024rotated} & Swin-B & BERT & 68.21 & 42.32 & 46.64 & 25.87 & 24.13 & 14.20 & 13.69 & 6.77 & 3.25 & 1.27 & 74.40 & 68.31 & 54.24 & 42.63 \\
			CrossVLT\cite{cho2023cross} & Swin-B & BERT & 67.52	& 41.94 &	43.85 & 25.43 & 25.99 & 15.19 & 14.62 & 3.71 & 1.87 & 1.76 & 76.12 & 69.73 & 55.27 & 42.81 \\
			RIS-DMMI\cite{hu2023beyond} & Swin-B & BERT & 86.17 & 63.89 & 74.71 & 44.30 & 38.05 & \textcolor{blue}{\textbf{19.81}} & 18.10 & 6.49 & 3.25 & 1.00 & 74.02 & 68.58 & 65.72 & 52.15 \\
			CARIS\cite{liu2023caris} & Swin-B & BERT & 68.45 & 45.40 & 47.10 & 27.19 & 25.52 & 15.08 & 14.62 & 7.87 & 3.71 & 1.98 & 75.79 & 69.74 & 54.30 & 42.66 \\
			robust-ref-seg\cite{wu2024towards} & Swin-B & BERT & 81.67 & 50.25 & 52.44 & 28.01 & 30.86 & 17.83 & 17.17 & 9.19 & \textcolor{blue}{\textbf{5.80}} & \textcolor{blue}{\textbf{2.48}} & 77.74 & 71.13 & 60.44 & 47.12 \\
			LGCE\cite{yuan2024rrsis} & Swin-B & BERT & 79.81	& 50.19 & 54.29	& 28.62 & 29.70 & 17.17 & 15.31 & 9.36 & 5.10 & 2.15 & 78.24 & 71.59 & 60.66 & 46.57
			\\ 
			LAVT\cite{yang2022lavt} & Swin-B & BERT & 80.97	& 51.84 & 58.70 & 30.27 & 31.09 & 17.34 & 15.55 & 9.52 & 4.64 & 2.09 & \textcolor{blue}{\textbf{78.50}} & 71.86 & 61.53 & 47.40
			\\ 
			\hline
			CroBIM (Ours) & Swin-B & BERT & \textcolor{blue}{\textbf{87.24}} & \textcolor{blue}{\textbf{64.83}} & \textcolor{blue}{\textbf{75.17}} & \textcolor{blue}{\textbf{44.41}} & \textcolor{blue}{\textbf{44.78}} & 17.28 & \textcolor{blue}{\textbf{19.03}} & \textcolor{blue}{\textbf{9.69}} & \textcolor{red}{\textbf{6.26}} & 2.20 &  \textcolor{red}{\textbf{78.85}} & \textcolor{blue}{\textbf{72.30}} & \textcolor{blue}{\textbf{65.79}} & \textcolor{blue}{\textbf{52.69}} \\	
			CroBIM (Ours) & ConvNeXt-B & BERT & \textcolor{red}{\textbf{93.04}} & \textcolor{red}{\textbf{75.89}} & \textcolor{red}{\textbf{87.70}} & \textcolor{red}{\textbf{61.42}} & \textcolor{red}{\textbf{66.13}} & \textcolor{red}{\textbf{34.07}} & \textcolor{red}{\textbf{26.91}} & \textcolor{red}{\textbf{12.99}} & \textcolor{blue}{\textbf{5.80}} & \textcolor{red}{\textbf{2.75}} & 77.95 & \textcolor{red}{\textbf{72.33}} & \textcolor{red}{\textbf{71.93}} & \textcolor{red}{\textbf{59.77}} \\	
			\hline
		\end{tabular}
	}
\end{table*}

\begin{figure*}[tbp]
	\centering
	\includegraphics[width=1\linewidth]{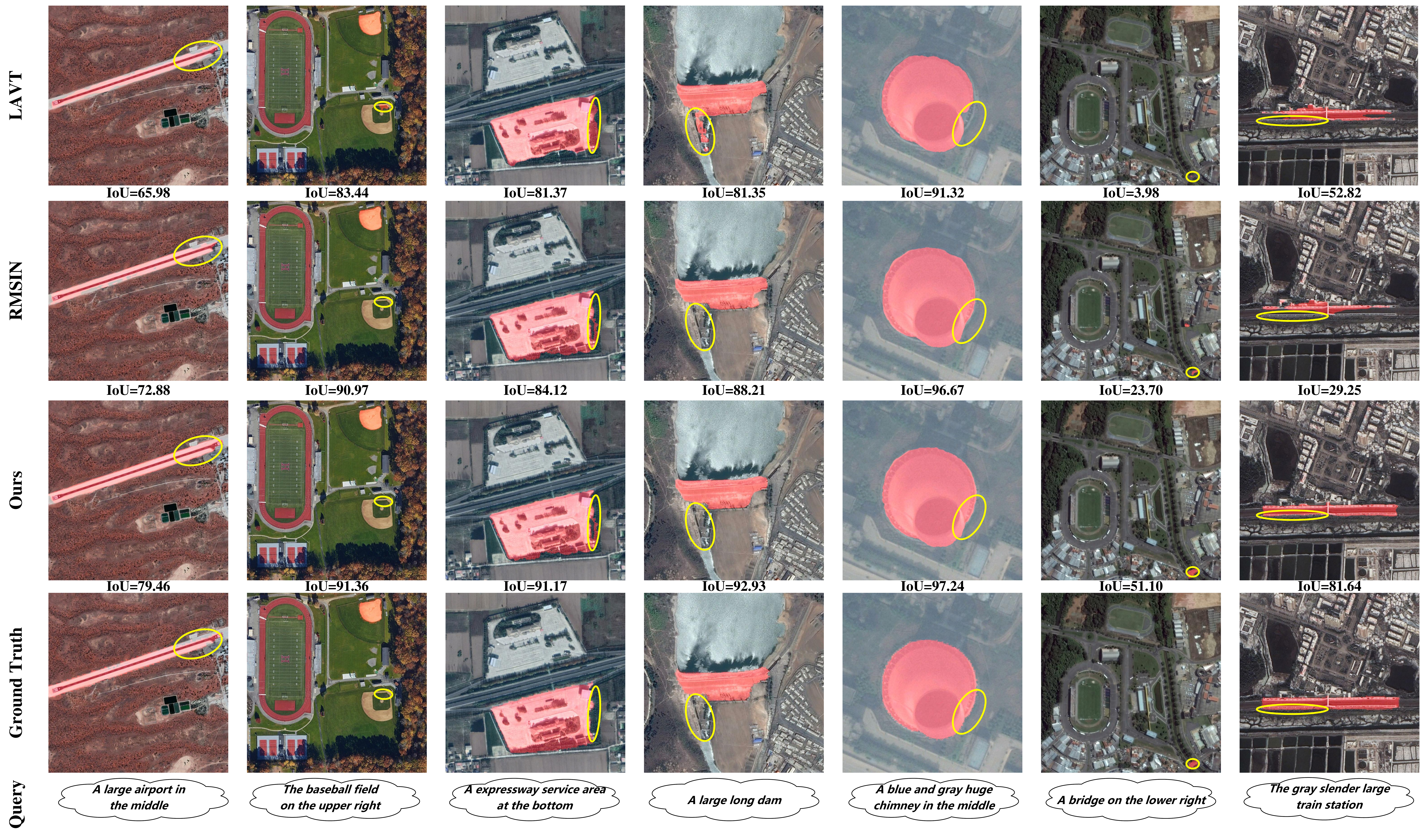}
	\caption{Visualization of segmentation results for CroBIM and comparison methods on the RRSIS-D dataset test set, with yellow circles highlighting areas of improved performance and corresponding IoU scores displayed.}
	\label{vis_rrsisd}
	
	\vspace{1em}  
	
	\includegraphics[width=1\linewidth]{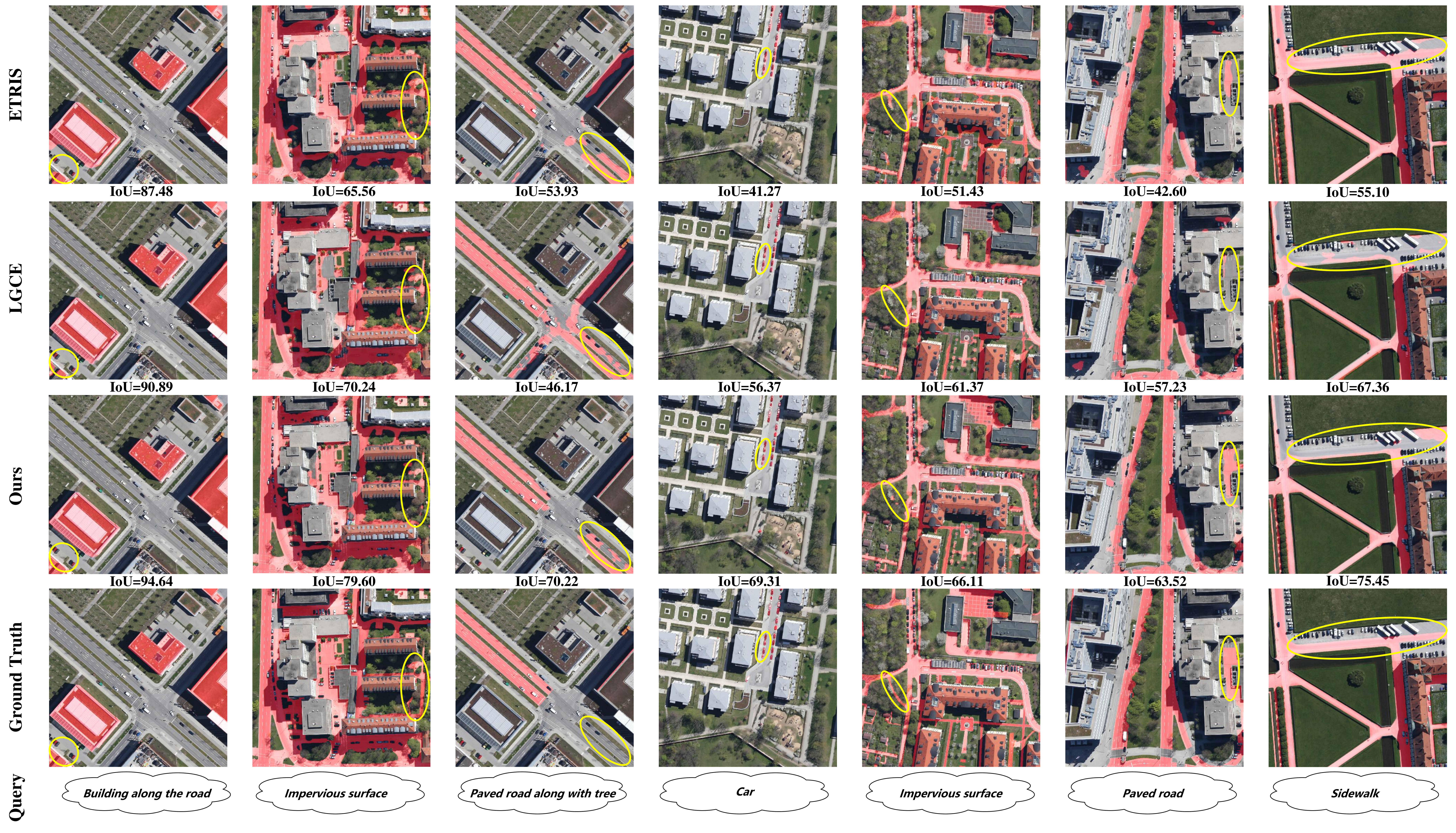}
	\caption{Visualization of segmentation results for CroBIM and comparison methods on the RefSegRS dataset test set, with yellow circles highlighting areas of improved performance and corresponding IoU scores displayed.}
	\label{vis_refsegrs}
\end{figure*}

%
%
\begin{table*}[htbp]
	\centering
	\caption{Comparison with state-of-the-art methods on the RISBench dataset. Optimal and sub-optimal performance in each metric are marked by \textcolor{red}{\textbf{red}} and \textcolor{blue}{\textbf{blue}}.}
	\label{risbench_comparison}
	\renewcommand\arraystretch{1.5}
	\fontsize{8}{12}\selectfont
	\resizebox{\textwidth}{!}{
		\begin{tabular}{l|c|c|c|c|c|c|c|c|c|c|c|c|c|c|c|c}
			\cmidrule{1-17} 
			\multirow{2}{*}{Method} & \multirow{2}{*}{Visual Encoder} & \multirow{2}{*}{Text Encoder} & \multicolumn{2}{c|}{Pr@0.5} & \multicolumn{2}{c|}{Pr@0.6} & \multicolumn{2}{c|}{Pr@0.7} & \multicolumn{2}{c|}{Pr@0.8} & \multicolumn{2}{c|}{Pr@0.9} & \multicolumn{2}{c|}{oIoU} & \multicolumn{2}{c}{mIoU} \\ \cline{4-17}  
			& & & Val & Test & Val & Test & Val & Test & Val & Test & Val & Test & Val & Test & Val & Test \\ 
			\cmidrule{1-17} 
			RRN\cite{li2018referring} & ResNet-101 & LSTM & 54.62 & 55.04 & 46.88 & 47.31 & 39.57 & 39.86 & 32.64 & 32.58 & 11.57 & 13.24 & 47.28 & 49.67 & 42.65 & 43.18 \\
			LSCM\cite{hui2020linguistic} & ResNet-101 & LSTM  & 55.87 & 55.26 & 47.24 & 47.14 & 40.22 & 40.10 & 33.55 & 33.29 & 12.78 & 13.91 & 47.99 & 50.08 & 43.21 & 43.69 \\
			BRINet\cite{hu2020bi} & ResNet-101 & LSTM & 52.11 & 52.87 & 45.17 & 45.39 & 37.98 & 38.64 & 30.88 & 30.79 & 10.28 & 11.86 & 46.27 & 48.73 & 41.54 & 42.91 \\
			MAttNet\cite{yu2018mattnet} & ResNet-101 & LSTM & 56.77 & 56.83 & 48.51 & 48.02 & 41.53 & 41.75 & 34.33 & 34.18 & 13.84 & 15.26 & 48.66 & 51.24 & 44.28 & 45.71 \\
			CMPC\cite{huang2020referring} & ResNet-101 & LSTM & 54.89 & 55.17 & 47.77 & 47.84 & 40.38 & 40.28 & 32.89 & 32.87 & 12.63 & 14.55 & 47.59 & 50.24 & 42.83 & 43.82\\
			CMPC+\cite{liu2021cross} & ResNet-101 & LSTM & 57.84 & 58.02 & 49.24 & 49.00 & 42.34 & 42.53 & 35.77 & 35.26 & 14.55 & 17.88 & 50.29 & 53.98 & 45.81 & 46.73 \\	
			ETRIS\cite{xu2023bridging} & ResNet-101 & CLIP & 59.87 & 60.98 & 49.91 & 51.88 & 35.88 & 39.87 & 20.10 & 24.49 & 8.54 & 11.18 & 64.09 & 67.61 & 51.13 & 53.06 \\
			CRIS\cite{wang2022cris} & ResNet-101 & CLIP & 63.42 & 63.67 & 54.32 & 55.73 & 41.15 & 44.42 & 24.66 & 28.80 & 10.27 & 13.27 & 66.26 & 69.11 & 53.64 & 55.18 \\
			LAVT\cite{yang2022lavt} & Swin-B & BERT & 68.27 & 69.40 & 62.71 & 63.66 & 54.46 & 56.10 & 43.13 & 44.95 & 21.61 & 25.21 & 69.39 & 74.15 & 60.45 & 61.93 \\
			RMSIN\cite{liu2024rotated} & Swin-B & BERT & 70.05 & 71.01 & 64.64 & 65.46 & 56.37 & 57.69 & 44.14 & 45.50 & 21.40 & 25.92 & 69.51 & 74.09 & 61.78 & 63.07 \\
			LGCE\cite{yuan2024rrsis} & Swin-B & BERT & 68.20 & 69.64 & 62.91 & 64.07 & 55.01 & 56.26 & 43.38 & 44.92 & 21.58 & 25.74 & 68.81 & 73.87 & 60.44 & 62.13 \\ 
			CrossVLT\cite{cho2023cross} & Swin-B & BERT & 70.05 & 70.62 & 64.29 & 65.05 & 56.97 & 57.40 & 44.49 & 45.80 & 21.47 & 26.10 & 69.77 & 74.33 & 61.54 & 62.84 \\ 
			CARIS\cite{liu2023caris} & Swin-B & BERT & 73.46 & 73.94 & 68.51 & 68.93 & 60.92 & 62.08 & 48.47 & 50.31 & 24.98 & \textcolor{blue}{\textbf{29.08}} & \textcolor{blue}{\textbf{70.55}} & \textcolor{red}{\textbf{75.10}} & 64.40 & 65.79 \\
			RIS-DMMI\cite{hu2023beyond} & Swin-B & BERT & 71.27 & 72.05 & 66.02 & 66.48 & 58.22 & 59.07 & 45.57 & 47.16 & 22.43 & 26.57 & \textcolor{red}{\textbf{70.58}} & \textcolor{blue}{\textbf{74.82}} & 62.62 & 63.93 \\
			robust-ref-seg\cite{wu2024towards} & Swin-B & BERT & 67.42 & 69.15 & 61.72 & 63.24 & 53.64 & 55.33 & 40.71 & 43.27 & 19.43 & 24.20 & 69.50 & 74.23 & 59.37 & 61.25 \\
			\hline
			CroBIM (Ours) & Swin-B & BERT & \textcolor{blue}{\textbf{76.59}} & \textcolor{blue}{\textbf{75.75}} & \textcolor{blue}{\textbf{71.73}} & \textcolor{blue}{\textbf{70.34}} & \textcolor{blue}{\textbf{64.32}} & \textcolor{blue}{\textbf{63.12}} & \textcolor{blue}{\textbf{53.18}} & \textcolor{blue}{\textbf{51.12}} & \textcolor{blue}{\textbf{28.53}} & 28.45 & 69.08 & 73.61 & \textcolor{blue}{\textbf{67.52}} & \textcolor{blue}{\textbf{67.32}} \\	
			CroBIM (Ours) & ConvNeXt-B & BERT & \textcolor{red}{\textbf{77.41}} & \textcolor{red}{\textbf{77.55}} & \textcolor{red}{\textbf{72.62}} & \textcolor{red}{\textbf{72.83}} & \textcolor{red}{\textbf{66.74}} & \textcolor{red}{\textbf{66.38}} & \textcolor{red}{\textbf{55.92}} & \textcolor{red}{\textbf{55.93}} & \textcolor{red}{\textbf{32.17}} & \textcolor{red}{\textbf{34.07}} & 69.12 & 73.04 & \textcolor{red}{\textbf{68.70}} & \textcolor{red}{\textbf{69.33}}  \\	
			\hline
		\end{tabular}
	}
\end{table*}

\subsection{Comparison with State-of-the-art Methods}

\subsubsection{RRSIS-D}

To evaluate the effectiveness of our proposed method, we conducted experiments on the RRSIS-D dataset. The comparison results are presented in Table.~\ref{rrsisd_comparison}. We compared CroBIM with several LSTM-based, CLIP-based, and BERT-based methods from classical to state-of-the-art, i.e., RRN\cite{li2018referring}, CSMA\cite{ye2019cross}, LSCM\cite{hui2020linguistic}, CMPC\cite{huang2020referring}, BRINet\cite{hu2020bi}, CMPC+\cite{liu2021cross}, BKINet\cite{ding2023bilateral}, ETRIS\cite{xu2023bridging}, CRIS\cite{wang2022cris}, LGCE\cite{yuan2024rrsis}, LAVT\cite{yang2022lavt}, RMSIN\cite{liu2024rotated}, CrossVLT\cite{cho2023cross}, RIS-DMMI\cite{hu2023beyond}, robust-ref-seg\cite{wu2024towards}, SLViT\cite{ouyang2023slvit}, CARIS\cite{liu2023caris}, MAFN\cite{shi2025multimodal}. Among the compared methods, LGCE, RMSIN, and MAFN are the only three approaches specifically designed for the RRSIS task in the remote sensing domain. All other methods included in our comparison are SOTA algorithms from the computer vision domain. Although not specifically tailored for remote sensing, these methods represent the current best practices in referring image segmentation and have been widely recognized within the computer vision community.

It can be observed that LSTM-based methods generally perform worse than CLIP-based and BERT-based methods. This is because, in the context of referring image segmentation, where the textual descriptions may require understanding of nuanced spatial relationships and contextual cues, LSTMs may struggle to fully capture the necessary semantics and syntactic variations due to their sequential processing nature. The combination of a Swin Transformer-based visual encoder and a BERT-based text encoder has become the mainstream solution for current referring image segmentation tasks.

Among all the compared methods, our CroBIM utilizes two different visual encoders (Swin-B and ConvNeXt-B) and a text encoder (BERT), achieving optimal or sub-optimal performance across multiple metrics. Notably, CroBIM with ConvNeXt-B as the visual encoder demonstrates outstanding performance in the majority of metrics, particularly achieving the best results in Pr@0.5, Pr@0.6, Pr@0.7, Pr@0.8, Pr@0.9, and mIoU. Specifically, CroBIM with ConvNeXt-B achieved 74.94\% (Val) and 74.58\% (Test) for Pr@0.5, and 67.64\% (Val) and 67.57\% (Test) for Pr@0.6, significantly outperforming other methods. Meanwhile, CroBIM with Swin-B achieved 74.20\% (Val) and 75.00\% (Test) for Pr@0.5, and 66.15\% (Val) and 66.32\% (Test) for Pr@0.6, also demonstrating strong performance. Notably, CroBIM (ConvNeXt-B) achieved higest mIoU scores of 65.05\% (Val) and 64.46\% (Test), which are 2.17\% (Val) and 2.34\% (Test) higher than the closest competing method, CARIS. Since the proposed CroBIM focuses on enhancing the model's ability to achieve precise segmentation of target categories, especially in complex remote sensing backgrounds, the model may attain higher local accuracy when handling challenging and less frequent target categories. This emphasis, however, may lead to suboptimal performance in oIoU. We also conduct a qualitative comparison between our model and the baseline to offer a comprehensive understanding of the results, as demonstrated in Fig.~\ref{vis_rrsisd}.

\begin{figure*}[tbp]
	\begin{center}
		\centerline{\includegraphics[width=0.92\linewidth]{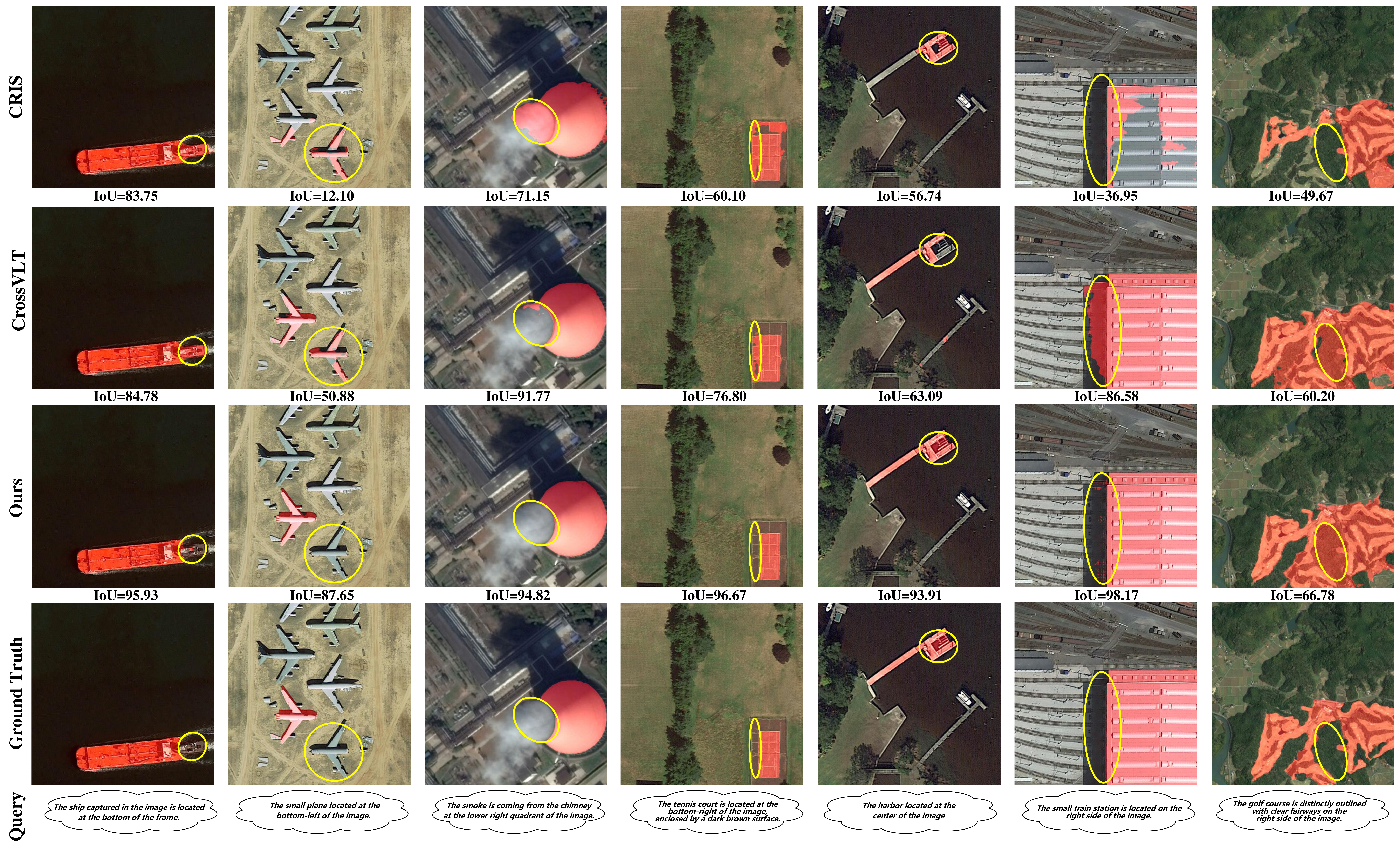}}
		\caption{Visualization of segmentation results for CroBIM and comparison methods on the RisBench dataset test set, with yellow circles highlighting areas of improved performance and corresponding IoU scores displayed.}\label{vis_risbench}
	\end{center}
\end{figure*}

\subsubsection{RefSegRS}

We further conduct experiments on the RefSegRS dataset to validate the superiority of our proposed framework, and the performance on both the validation and test sets is reported in Table.~\ref{refsegrs_comparison}. First, in terms of the Pr@0.5 and Pr@0.6 metrics, CroBIM consistently outperforms other state-of-the-art methods on both the validation and test sets. Notably, CroBIM with ConvNeXt-B as the visual encoder achieves the highest accuracy of 75.89\% on the test set for Pr@0.5, significantly surpassing the second-best method, RIS-DMMI, which attained 63.89\%. This result demonstrates that CroBIM effectively segments the target regions under lower overlap thresholds. Additionally, CroBIM continues to exhibit a marked advantage in the Pr@0.6 metric, achieving an accuracy of 61.42\% on the test set, which is substantially higher than RIS-DMMI's 44.30\%. As the overlap threshold increases, CroBIM maintains robust performance. Specifically, at the more stringent Pr@0.9 threshold, CroBIM attains an accuracy of 6.26\% on the test set, further confirming its effectiveness in high-precision segmentation tasks.

CroBIM also demonstrates superior performance in two key metrics: oIoU and mIoU. For oIoU, CroBIM achieves test scores of 72.33\% (ConvNeXt-B) and 72.30\% (Swin-B), ranking first and second respectively, significantly outperforming competing methods such as LAVT and RIS-DMMI. Similarly, in terms of mIoU, CroBIM achieves 59.77\% (ConvNeXt-B) and 52.69\% (Swin-B) on the test set, once again outperforming all other methods. These results highlight that CroBIM not only excels at lower overlap thresholds but also delivers significant improvements in overall segmentation accuracy. Fig.~\ref{vis_refsegrs} presents a qualitative analysis contrasting our model with the comparison methods, providing insights into the performance differences.

\subsubsection{RisBench}

In our constructed RISBench dataset, the proposed CroBIM model also demonstrates significant performance advantages over competing methods, achieving either the best or second-best results across multiple evaluation metrics, as shown in Table.~\ref{risbench_comparison}. Notably, CroBIM outperforms the current state-of-the-art models on both the validation and test sets across various threshold levels, including Pr@0.5, Pr@0.6, Pr@0.7, Pr@0.8, and Pr@0.9. These results further validate the effectiveness and robustness of our model. Specifically, the CroBIM model, utilizing the ConvNeXt-B visual encoder and BERT text encoder, achieves the highest test set precision at Pr@0.5, reaching 77.55\%, outperforming all other methods. Additionally, at Pr@0.6, Pr@0.7, and Pr@0.8, the model attains precisions of 72.83\%, 66.38\%, and 55.93\%, respectively, significantly surpassing existing methods. Particularly, at the higher threshold of Pr@0.9, CroBIM leads other methods with a precision of 34.07\%, demonstrating the model's strong capability to handle more complex scenarios that demand higher precision.

In contrast, although models such as RIS-DMMI and CARIS exhibit competitive performance on certain metrics—CARIS achieves 75.10\% on oIoU and RIS-DMMI reaches 70.58\%—these models fail to maintain consistent performance under higher precision thresholds. CroBIM, on the other hand, exhibits competitive results across both comprehensive metrics, oIoU and mIoU. The ConvNeXt-B variant of CroBIM achieves a test set mIoU of 69.33\%, surpassing all comparison methods. This indicates that our model excels not only in individual precision metrics but also in overall segmentation accuracy.

Compared to other models utilizing the Swin-B and BERT combinations, CroBIM significantly enhances vision-language alignment between visual and textual features through improvements in feature extraction. By leveraging an innovative cross-modal bidirectional interaction mechanism, our model more effectively captures key information in the image that corresponds to the textual description. This capability is particularly beneficial in remote sensing imagery where targets are complex and the background is often noisy. Even under such challenging conditions, CroBIM consistently achieves excellent segmentation results. The visualization results in Fig.~\ref{vis_risbench} further highlight the superiority of our approach.

Besides, the experimental results demonstrated that our proposed CroBIM framework not only outperformed the existing RRSIS-specific methods (LGCE, RMSIN and MAFN) but also surpassed the adapted SOTA methods from computer vision. This highlighted the effectiveness and robustness of our approach in handling the unique challenges of remote sensing imagery, such as complex spatial relationships and diverse object scales.

\section{Ablation Study}
\label{section:Ablation}

To assess the effectiveness of our designs within the CroBIM framework, we perform comprehensive experiments using our constructed RISBench dataset and present the quantitative outcomes obtained from the test set analysis.

\subsection{Effectiveness of Each Components}

\begin{figure}[tbp]
	\begin{center}
		\centerline{\includegraphics[width=1\linewidth]{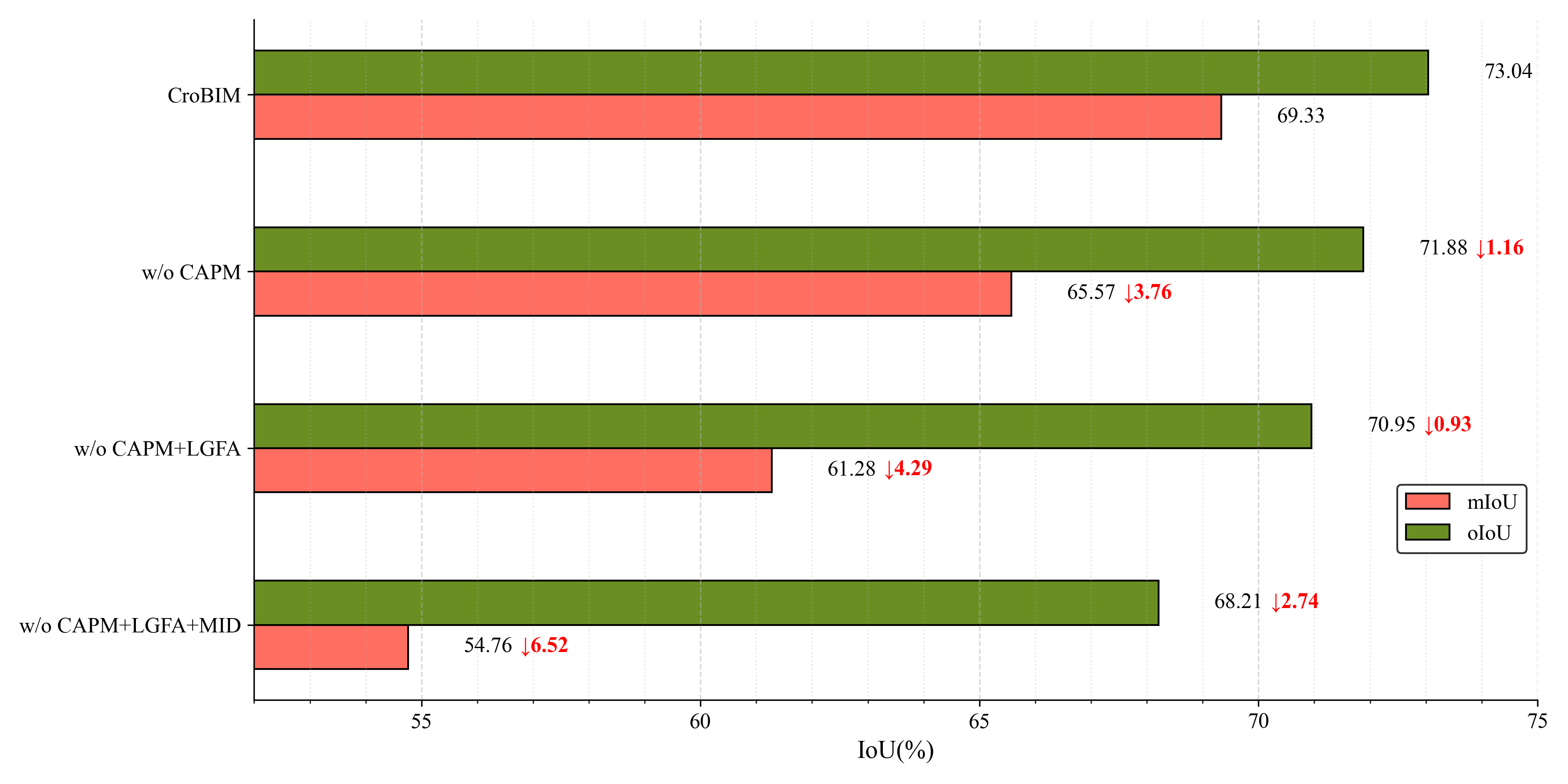}}
		
		\caption{Performance comparison of CroBIM and its ablated variants (w/o CAPM, LGFA, and MID) on the RISBench test set.}\label{ablation}
	\end{center}
\end{figure}

To evaluate the effectiveness of the three key components of CroBIM—CAPM, LGFA, and MID—we conducted ablation experiments on the RISBench test dataset. The experimental results are presented in Fig.~\ref{ablation}.

When the CAPM module is removed from the complete CroBIM framework, the mIoU drops from 69.33\% to 65.57\%, representing a significant performance reduction of 3.76\%. Similarly, the oIoU decreases from 73.04\% to 71.88\%. These results underscore the critical role of CAPM in integrating multi-scale spatial context into linguistic features, thereby enabling precise localization of target objects in remote sensing images. In the absence of CAPM, the model struggles to capture the spatial positional relationships described in the referring expressions, leading to a noticeable decline in segmentation accuracy.

When the LGFA module is further removed, the performance degradation becomes more pronounced. The mIoU decreases to 61.28\% (a reduction of 4.29\%) and the oIoU drops to 70.95\% (a reduction of 0.93\%) compared to the prior configuration. This highlights the pivotal role of LGFA in aggregating visual and linguistic features across multiple scales and compensating for attention deficits. Without LGFA, the model loses its ability to effectively handle objects of varying scales, which is a fundamental challenge in remote sensing scenarios.

Finally, replacing the MID module with a simple decoder\cite{yang2022lavt} results in mIoU and oIoU dropping further to 54.76\% and 68.21\%, respectively. These findings clearly demonstrate the importance of MID in ensuring precise alignment between visual and linguistic features through iterative bidirectional refinement. The absence of MID significantly impairs the model's ability to resolve the complex cross-modal relationships required for accurate segmentation, particularly in remote sensing datasets characterized by diverse geospatial relationships and linguistically rich referring expressions.

In summary, the above results conclusively demonstrate that all three components are indispensable for achieving precise segmentation in complex scenarios. Each module contributes uniquely to the overall performance of CroBIM, ensuring its robustness and effectiveness in addressing the challenges inherent in RRSIS tasks.

\begin{figure*}[htbp]
	\begin{center}
		\centerline{\includegraphics[width=1\linewidth]{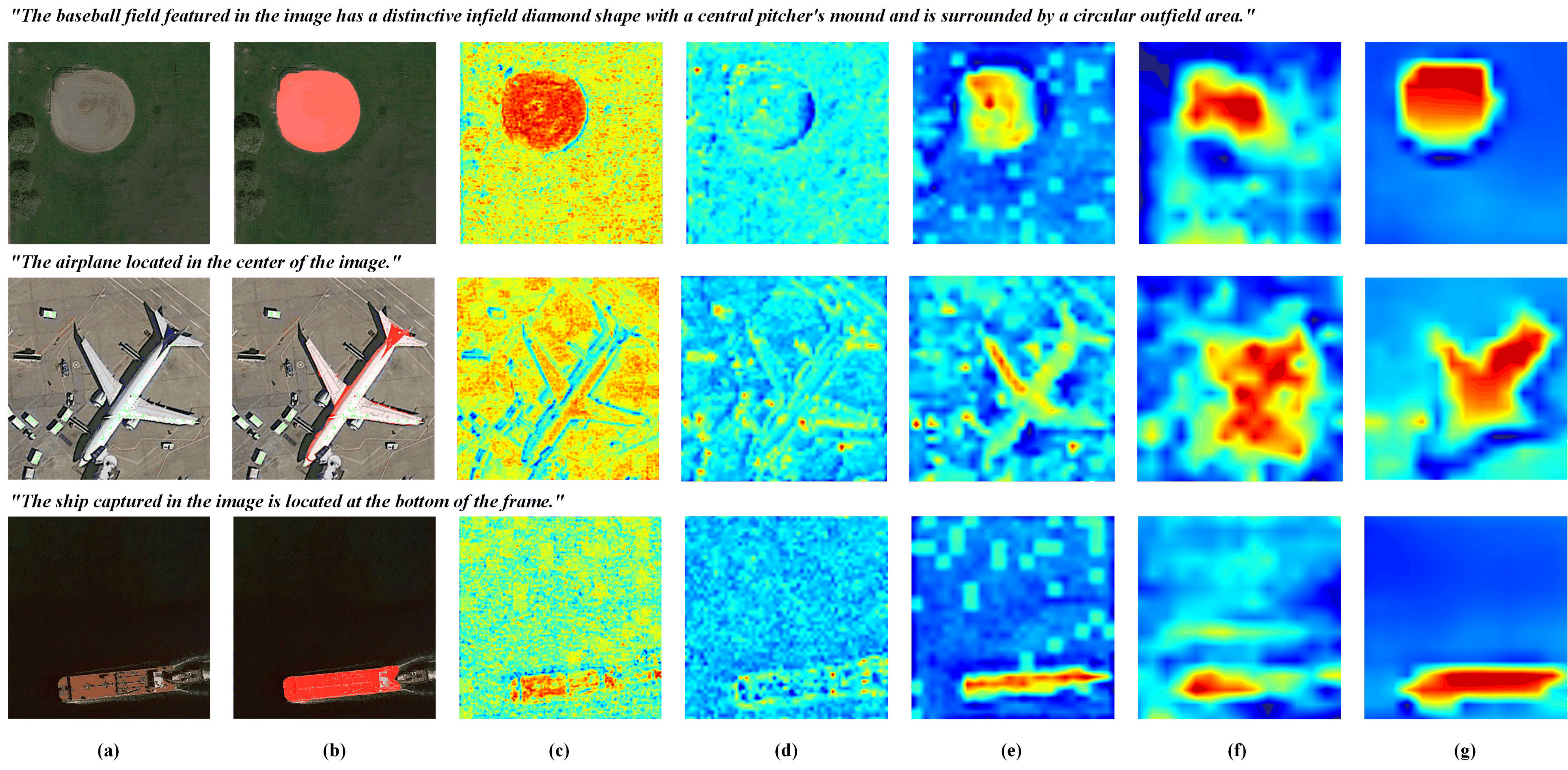}}
		\caption{Attention map visualization from different stages in CroBIM. (a) input image, (b) ground truth, (c)-(f) attentions maps of S1-S4 stages in the encoder, (g) attentions maps of the decoder. }\label{vis_attn}
	\end{center}
\end{figure*}
%

\begin{table}[tbp]
	\centering
	\scriptsize 
	\caption{Ablation studies on options design of CAPM.}
	\label{ab_capm}
	\renewcommand{\arraystretch}{1.4}
	\setlength{\tabcolsep}{3.8pt} 
	\begin{tabular}{l|c|c|c|c|c}
		\toprule
		\textbf{Option} & \textbf{Pr@0.5} & \textbf{Pr@0.7} & \textbf{Pr@0.9} & \textbf{oIoU} & \textbf{mIoU} \\
		\midrule
		\multicolumn{6}{l}{\textbf{(a) Combination of visual features at different scales}} \\ \hline
		\{$V_1$, $V_2$, $V_3$\}            & 77.22 & 66.28 & 33.96 & 72.95 & 69.04 \\
		\{$V_1$, $V_2$, $V_4$\}            & 77.15 & 66.25 & 33.82 & 72.88 & 69.11 \\
		\{$V_2$, $V_3$, $V_4$\}            & 77.26 & 66.24 & 33.89 & \textbf{73.16} & 69.15 \\
		\{$V_1$, $V_2$, $V_3$, $V_4$\}     & \textbf{77.55} & \textbf{66.38} & \textbf{34.07} & 73.04 & \textbf{69.33} \\
		\midrule
		\multicolumn{6}{l}{\textbf{(b) Design of Pooling}} \\ \hline
		Max Pool                 & 76.38 & 65.04 & 33.52 & 72.47 & 68.67 \\
		Average Pool             & 77.23 & 66.15 & 33.61 & 72.84 & 69.10 \\
		Adaptive Average Pool    & \textbf{77.55} & \textbf{66.38} & \textbf{34.07} & \textbf{73.04} & \textbf{69.33} \\
		\bottomrule
	\end{tabular}
\end{table}

\subsection{Design of CAPM}

To further validate the effectiveness of the CAPM module, we conduct a detailed analysis of its context-aware prompt design. First, we performe ablation studies on the combinations of visual features at different scales, with the results presented in Table.~\ref{ab_capm}(a). As can be observed, the combination incorporating all four visual features, ${V_1, V_2, V_3, V_4}$, achieves the best performance in terms of prediction accuracy (Pr@0.5, Pr@0.7, Pr@0.9) and mIoU. This demonstrates that integrating visual features from multiple scales significantly enhances the model’s prediction accuracy and overall segmentation quality. The improved performance suggests that the comprehensive utilization of features across different levels of granularity contributes to a more robust and precise representation, thus improving the model's ability to capture complex visual cues.

We further investigate the impact of different pooling strategies in the dimensionality reduction of multi-scale visual features, as shown in Table.~\ref{ab_capm}(b). It can be observed that, compared to max pooling and average pooling, adaptive average pooling consistently achieves superior performance across all evaluation metrics. These results suggest that the adaptive mechanism is more effective in flexibly integrating features across different scales, thereby preserving more information that is crucial for accurate predictions. Moreover, by dynamically adjusting the weighting of features based on their respective scales, adaptive average pooling enables a more comprehensive retention of both global and local information during the dimensionality reduction process, ultimately enhancing segmentation performance.

\subsection{Design of LGFA}

We further validated the effectiveness of attention deficit compensation within the LGFA module. The ablation study results presented in Table \ref{ab_lgfa} demonstrate the efficacy of applying attention deficit compensation at different stages (S1-S4) in improving the aggregation and alignment of visual and linguistic features. As attention deficit compensation is progressively applied from a single stage (S4) to multiple stages, a significant performance improvement across all metrics can be observed. Notably, when attention deficit compensation is applied at all stages (S1-S4), the model achieves the best segmentation results: Pr@0.5 of 77.55\%, Pr@0.7 of 66.38\%, and mIoU of 69.33\%. The most prominent gains are observed in Pr@0.5 and oIoU, indicating that the complete attention deficit compensation strategy enhances the precision of feature alignment and overall segmentation accuracy. These findings underscore the importance of applying attention deficit compensation at multiple stages, as it leads to more robust integration of visual and linguistic features, thereby improving the segmentation performance.

%

\begin{table}[tbp]
	\centering
	\scriptsize
	\caption{Ablation studies on attention deficit compensation of LGFA.}
	\label{ab_lgfa}
	\renewcommand{\arraystretch}{1.4}
	\setlength{\tabcolsep}{4pt}
	\begin{tabular}{cccc|c|c|c|c|c}
		\toprule
		\textbf{S1} & \textbf{S2} & \textbf{S3} & \textbf{S4} 
		& \textbf{Pr@0.5} & \textbf{Pr@0.7} & \textbf{Pr@0.9} & \textbf{oIoU} & \textbf{mIoU} \\
		\midrule
		&  &  & \checkmark                & 76.64 & 65.72 & 33.61 & 72.18 & 68.55 \\
		&  & \checkmark & \checkmark       & 77.01 & 65.93 & 33.85 & 72.61 & 68.78 \\
		& \checkmark & \checkmark & \checkmark & 77.14 & 66.21 & \textbf{34.12} & 72.87 & 69.02 \\
		\checkmark & \checkmark & \checkmark & \checkmark & \textbf{77.55} & \textbf{66.38} & 34.07 & \textbf{73.04} & \textbf{69.33} \\
		\bottomrule
	\end{tabular}
\end{table}

\vspace{1em}

\begin{table}[tbp]
	\centering
	\scriptsize
	\caption{Ablation studies on attention mechanisms of MID.}
	\label{ab_mid}
	\renewcommand{\arraystretch}{1.4}
	\setlength{\tabcolsep}{5pt}
	\begin{tabular}{l|c|c|c|c|c}
		\toprule
		\textbf{Attention} & \textbf{Pr@0.5} & \textbf{Pr@0.7} & \textbf{Pr@0.9} & \textbf{oIoU} & \textbf{mIoU} \\
		\midrule
		PWAM \cite{yang2022lavt}        & 72.64 & 63.87 & 30.59 & 70.28 & 65.54 \\
		WPA \cite{zhang2022coupalign}   & 75.83 & 64.87 & 32.71 & 72.19 & 67.02 \\
		\textbf{CBAM (ours)}            & \textbf{77.55} & \textbf{66.38} & \textbf{34.07} & \textbf{73.04} & \textbf{69.33} \\
		\bottomrule
	\end{tabular}
\end{table}

\subsection{Design of MID}
Moreover, we investigat the effectiveness of the cascaded bidirectional attention mechanism (CBAM) employed in our MID, and compared it with two existing approaches: the unidirectional visual-to-language attention module PWAM\cite{yang2022lavt}, and the parallel bidirectional attention WPA\cite{zhang2022coupalign}. The comparative results are presented in Table.~\ref{ab_mid}. As observed, the unidirectional attention mechanism in PWAM yielded the lowest segmentation performance. This underperformance can be attributed to its reliance on unidirectional interactions, which fail to adequately capture the joint representation of cross-modal visual-linguistic features.

In contrast, our cascaded bidirectional attention mechanism achieved the best performance, outperforming PWAM and WPA by 3.79\% and 2.31\% in terms of mIoU, respectively. These results demonstrate that the cascaded bidirectional attention effectively enables deep interaction between linguistic features and multi-scale visual contexts, fostering vision-language alignment in a more efficient manner. This enhanced interaction not only improves the quality of feature representation but also significantly boosts segmentation performance, highlighting the superiority of our proposed approach in addressing the challenges of visual-linguistic cross-modal tasks.

\begin{figure}[tbp]
	\begin{center}
		\centerline{\includegraphics[width=1\linewidth]{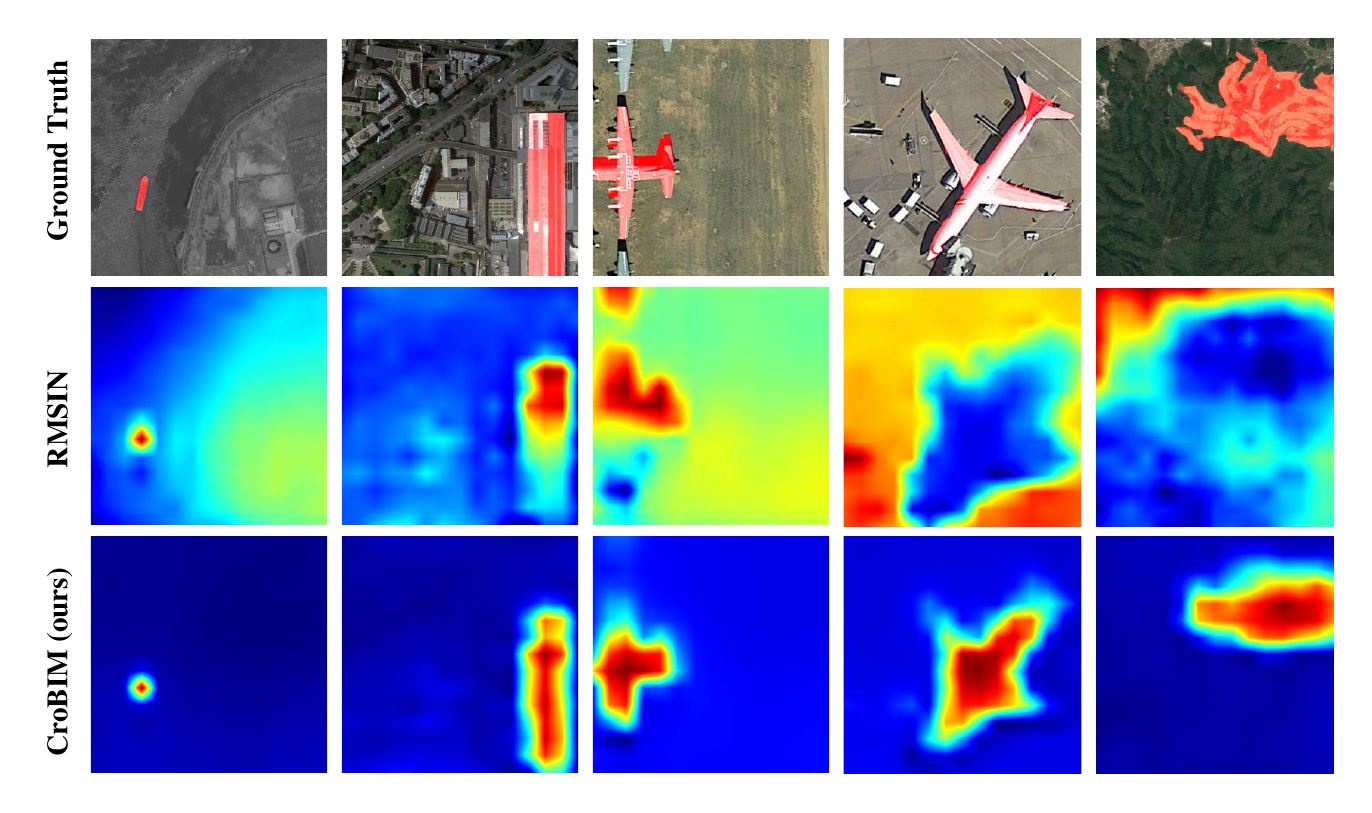}}
		\caption{Comparison of attention maps between CroBIM and RMSIN at the decoder stage.}\label{attn_compare}
		\vspace{-20pt}
	\end{center}
\end{figure}

\subsection{Qualitative Results}

In Fig.~\ref{vis_attn}, we visualize the attention maps at various stages of the model to analyze CroBIM's vision-language alignment mechanism in greater depth. First, in the encoder's attention maps, we observe that as the network depth increases, CroBIM progressively focuses on more fine-grained target areas. The attention maps from the early stages reveal a broad feature capture, with attention distributed across the global information in the image. This is closely related to the high resolution and complex background of remote sensing imagery—at these early stages, the model must focus on the overall structure to establish a comprehensive understanding of the scene. As the network layers progress, attention increasingly narrows to target regions that are closely aligned with the textual description, indicating that the model effectively filters out background noise and captures the key features of the target. For instance, in remote sensing images with complex terrain or buildings, CroBIM is able to accurately localize critical objects such as buildings and roads in the mid-to-late layers. Furthermore, the attention maps in the decoder stage illustrate the model’s segmentation capability. In the decoder, the attention maps demonstrate a high degree of refinement, with the model exhibiting more precise attention to the edges and finer details of the target regions. This focus further validates CroBIM’s strong performance in processing high-resolution remote sensing images, particularly in tasks requiring precise target segmentation. The decoder’s attention maps clearly show that the model can accurately delineate the boundaries between the target and background, especially in areas where textures or colors are highly similar.

\begin{figure}[tbp]
	\begin{center}
		\centerline{\includegraphics[width=1\linewidth]{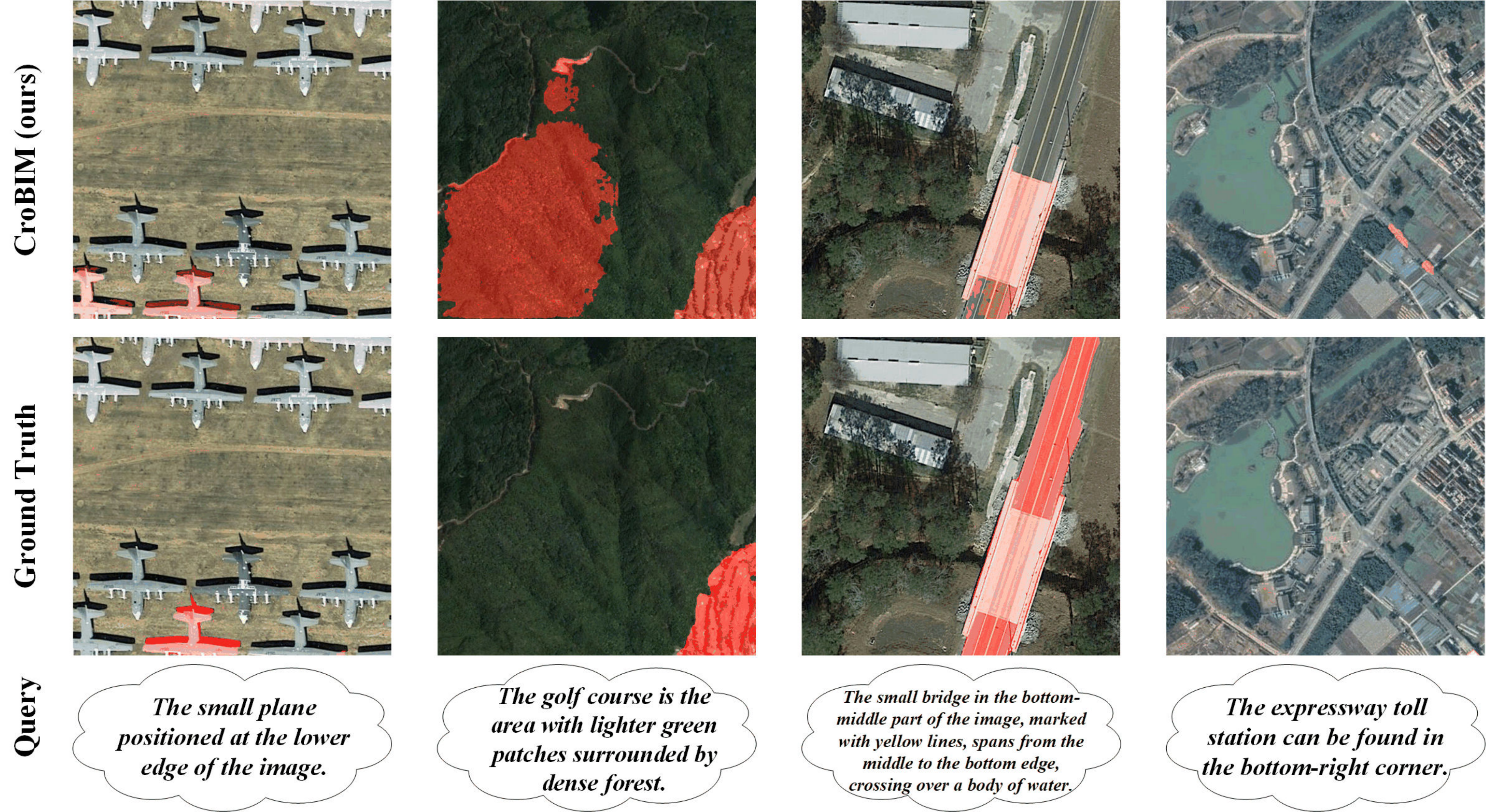}}
		\caption{Failure cases of our proposed CroBIM on the RisBench test set. From left to right, the four failure types are semantic ambiguity, visual similarity, imprecise annotations, and boundary targets.}\label{failure}
		\vspace{-20pt}
	\end{center}
\end{figure}

To further demonstrate the superiority of CroBIM, we compared its attention maps with those of RMSIN at the final stage of the decoder, as shown in Fig.~\ref{attn_compare}. The comparison reveals that RMSIN often suffers from attention drift, failing to focus accurately on the target regions specified in the textual query. In contrast, CroBIM consistently maintains precise attention on the target objects, even in complex scenarios with challenging geospatial relationships or similar foreground-background appearances. This refined attention mechanism enables CroBIM to achieve high-precision segmentation by accurately delineating target boundaries and capturing fine details, particularly in regions with similar textures or colors.


Although our CroBIM effectively models the query expression and aligns visual features with textual embeddings, some failure cases still occur, as shown in Fig.~\ref{failure}. These failure cases can be categorized into four types: semantic ambiguity, visual similarity, imprecise annotations, and boundary targets. First, semantic ambiguity arises when the textual description corresponds to multiple objects within the image, leading the model to segment multiple instances instead of the intended target. Second, visual similarity occurs when the foreground object closely resembles the background in appearance, making it difficult for the model to distinguish between them. Third, imprecise annotations refer to errors introduced by inaccurate or inconsistent labeling, which can mislead the model during training and evaluation. Finally, boundary targets present challenges when the object is located near the image boundary, where limited contextual information may cause the model to misinterpret its position and shape, resulting in inaccurate segmentation.

\section{Conclusion}
\label{section:Conclusion}

In this paper, we introduce the CroBIM framework to address the challenge of referring image segmentation in remote sensing scenarios. By leveraging bidirectional visual-text feature interaction and alignment, CroBIM bridges the gap between visual perception and language understanding, facilitating precise target segmentation. Furthermore, we construct a large-scale benchmark dataset, RISBench, which encompasses a more extensive set of image-language-label triplets, richer attribute expressions, a broader range of spatial resolutions, and more detailed textual descriptions. Experimental results demonstrate that the proposed CroBIM framework outperforms state-of-the-art methods across three benchmark datasets, underscoring its efficacy and superiority.

The ability to precisely segment target objects in remote sensing images based on natural language descriptions not only enhances the accessibility of remote sensing data but also opens up new possibilities for interdisciplinary research and practical applications in fields such as disaster response, urban planning, and environmental monitoring. For future work, we aim to embed domain-specific knowledge\cite{wei2021finetuned,10485462} of remote sensing imagery into language models, such as sensor imaging theory, spatial correlations, and spectral characteristics of ground objects, to further enhance remote sensing data analysis and interpretation. Additionally, another promising research direction is the integration of textual information with remote sensing through geolocation. By incorporating non-traditional geographic data, such as geotagged social media posts\cite{zhu2022geoinformation} and newspaper articles, remote sensing data can be combined with complementary sources. This approach broadens the potential applications\cite{haberle2022can, kruspe2020cross, kruspe2021detection} of remote sensing visual-language foundation models.

\section*{Declaration of Competing Interest}

The authors declare that they have no known competing financial interests or personal relationships that could have appeared to influence the work reported in this paper. 

\section*{Acknowledgements}

The authors would like to thank the Editor, Associate Editor, and anonymous reviewers for their helpful comments and suggestions that
improved this paper. This work was supported by the Distinguished Young Scholars of Natural Science Foundation of China under Grant 62025107.

\bibliographystyle{isprs}
\bibliography{refs}

\end{document}